\documentclass[journal]{new-aiaa} 
\usepackage[utf8]{inputenc}
\usepackage{multirow}
\usepackage{float}

\usepackage{subcaption}
\usepackage{graphicx}
\usepackage{amsmath}
\usepackage[version=4]{mhchem}
\usepackage{siunitx}
\usepackage{longtable,tabularx}
\title{Compact and Intuitive Airfoil Parameterization through Physics-aware Variational Autoencoder}

\author{Yu-Eop Kang\footnote{Department of Aerospace Engineering, Seoul National University, kye72594@snu.ac.kr} and Dawoon Lee\footnote{Department of Aerospace Engineering, Seoul National University, downl3824@snu.ac.kr}}

\affil{Department of Aerospace Engineering, Seoul National University, Seoul, 08826, Republic of Korea}
\author{Kwanjung Yee\footnote{Institute of Advanced Aerospace Technology, Seoul National University, kjyee@snu.ac.kr}\footnote{Corresponding Author}}
\affil{Institute of Advanced Aerospace Technology, Seoul National University, Seoul, 08826, Republic of Korea}

\begin{document}

\maketitle

\begin{abstract}
Airfoil shape optimization plays a critical role in the design of high-performance aircraft. However, the high-dimensional nature of airfoil representation causes the challenging problem known as the "curse of dimensionality". To overcome this problem, numerous airfoil parameterization methods have been developed, which can be broadly classified as polynomial-based and data-driven approaches. Each of these methods has desirable characteristics such as flexibility, parsimony, feasibility, and intuitiveness, but a single approach that encompasses all of these attributes has yet to be found. For example, polynomial-based methods struggle to balance parsimony and flexibility, while data-driven methods lack in feasibility and intuitiveness. In recent years, generative models, such as generative adversarial networks and variational autoencoders, have shown promising potential in airfoil parameterization. However, these models still face challenges related to intuitiveness due to their black-box nature. To address this issue, we developed a novel airfoil parameterization method using physics-aware variational autoencoder. The proposed method not only explicitly separates the generation of thickness and camber distributions to produce smooth and non-intersecting airfoils, thereby improving feasibility, but it also directly aligns its latent dimensions with geometric features of the airfoil, significantly enhancing intuitiveness. Finally, extensive comparative studies were performed to demonstrate the effectiveness of our approach.
\end{abstract}

\newpage


\section{Introduction}
\lettrine{A}{irfoil} shape optimization (ASO) plays a central role in the design of high-performance wings, blades and propellers. This becomes a multi-faceted design challenge in three-dimensional wings, where each airfoil section encounters varying angles of attack and Mach numbers during operation, making ASO far more complex than a single-point problem \cite{cook2017robust}. This complexity is further amplified in rotor blade design, because the combination of the tangential rotational speed, the asymmetric free stream speed in forward flight, and the vortex-induced effect results in a variable relative speed within a single revolution \cite{massaro2012multi}. Therefore, to achieve optimal aerodynamic performance, it is important to explore a wide design space that encompasses multiple candidate designs suitable for different local flow conditions.

However, the high-dimensional nature of airfoil geometry representation poses the challenging problem of the "curse of dimensionality". It not only increases the complexity of data acquisition, requiring a significant amount of CFD evaluations, but also results in an ill-conditioned design space, making design space exploration intractable. Undoubtedly, the airfoil parameterization method - how to effectively represent its high-dimensional coordinates - can have a significant impact on the efficiency of ASO.

In recent decades, numerous comparative studies have been conducted to evaluate airfoil parameterization techniques, highlighting the properties that ideal methods should possess \cite{samareh2001survey, wu2003comparisons, song2004study, sripawadkul2010comparison, zhu2014intuitive, masters2017geometric}. From these, four desirable characteristics can be drawn: 1) \textit{\textbf{flexibility}} - the ability to encompass a wide range of design candidates, 2) \textit{\textbf{parsimony}} - the ability to represent the airfoil with the fewest design variables, 3) \textit{\textbf{feasibility}} - the assurance of generating smooth and non-intersecting airfoils that can be seamlessly integrated into the ASO process, and 4) \textit{\textbf{intuitiveness}} - the straightforward mapping between design variables and the airfoil's geometric features, which supports the direct imposition of physical constraints and promotes efficient knowledge exchange among research groups.

Over the years, various airfoil parameterization techniques have emerged to achieve the aforementioned desirable properties \cite{sobieczky1999parametric, kulfan2008universal, piegl1996nurbs, toal2010geometric, derksen2010bezier, lu2018improved}. Traditional airfoil parameterization techniques, such as the parameterized section (PARSEC) \cite{sobieczky1999parametric}, class/shape-function transformation (CST) \cite{kulfan2008universal}, B-splines/B\'ezier curve representation \cite{piegl1996nurbs}, B\'ezier-PARSEC \cite{derksen2010bezier}, and improved geometric parameterization (IGP) \cite{lu2018improved}, characterize airfoil geometries using polynomial bases by adjusting their coefficients. This polynomial-based representation has several advantages: First, it ensures the smoothness of the resulting airfoil. Second, it provides direct access to the geometry derivatives, making it easier to relate the coefficients to physically significant features, such as the leading edge radius, the position of the upper/lower crest, and maximum thickness/camber (as seen in PARSEC \cite{sobieczky1999parametric}, B\'ezier-PARSEC \cite{derksen2010bezier}, and IGP \cite{lu2018improved}). Finally, high flexibility can easily be achieved by increasing the polynomial degree (as in CST \cite{kulfan2008universal} and B-splines/B\'ezier curve representations \cite{piegl1996nurbs}).

While the benefits of traditional airfoil parameterization methods are clear, such techniques can pose challenges within the ASO process. Specifically, using a polynomial of higher order leads to an increase in the number of design variables and frequently leads to the generation of oscillating and self-intersecting surfaces. These not only reduce the efficiency of data preparation, but also slow down the convergence of the optimization. To mitigate these problems, designers typically reduce the number of design variables or select a reference airfoil and constrain the design space by setting the arbitrary bounding box. However, both approaches severely limit the flexibility of the parameterization methods and rely heavily on the user's prior knowledge.

To strike a balance between parsimony and flexibility, data-driven techniques can be used. Unlike traditional methods that pre-define the design space using a polynomial system (top-down approach), data-driven methods extract the significant features from historical airfoils, such as the UIUC airfoil database \cite{selig1996uiuc}, and treat them as a design variables (bottom-up approach). Since historical airfoil candidates are the results of multidisciplinary optimization efforts, they naturally tend to cluster in a narrower region than the design space defined by high-order polynomial representations \cite{bengio2013representation}. Consequently, the data-driven approaches are likely to yield more parsimonious design variables than the traditional polynomial-based approach, without sacrificing their flexibility.

One popular data-driven method is singular value decomposition (SVD) \cite{toal2010geometric}. As a dimensionality reduction method, SVD first extracts the low-dimensional modes from the given dataset through eigendecomposition, and then generate the design candidates using linear combination of eigenvectors. However, SVD-based methods can only represent the airfoil as a linear combination of the extracted modes, compromising the compactness of the airfoil parameterization. Moreover, it is observed that unfeasible intersecting airfoils are generated, which eventually degrades the optimization convergence \cite{chen2020airfoil}.

Recently, there have been several attempts to address these issues using deep neural network-based generative models such as generative adversarial network (GAN) \cite{goodfellow2020generative} and variational autoencoder (VAE) \cite{kingma2013auto}. Originally, both GANs and VAEs emerged as prominent models in the field of computer vision due to their ability to generate realistic samples from learned latent spaces. The GAN is based on game theory principles, setting up a min-max adversarial game between the generator and the discriminator networks. The goal is for the generator to produce samples that the discriminator cannot distinguish from real data. While GAN can generate extremely realistic samples, it suffers from the problem of unstable training and mode collapse \cite{arjovsky2017wasserstein, salimans2016improved}. On the other hand, VAE employs Kullback-Leibler divergence (KLD) loss to regularize the latent space, ensuring that it conforms to a prior distribution, typically a Gaussian distribution with a diagonal covariance matrix. Although VAEs show better training stability compared to GANs, they tend to over-smooth the output data, which is often considered a drawbacks, especially when reconstructing the high-resolution image \cite{razavi2019generating}.

Building on their fundamental success in computer vision, researchers have also begun to explore the applications of GANs and VAEs in the field of ASO \cite{yilmaz2020conditional, achour2020development, yonekura2021data, yang2023inverse}. For example, Achour et al. \cite{achour2020development} and Yilmax et al. \cite{yilmaz2020conditional} used conditional GAN (CGAN) for the inverse design of airfoils. They first treated the aerodynamic performance of airfoils as a label to train CGAN, and then used the trained model to generate candidate airfoils that were expected to achieve the target performance. Similar studies based on VAE have also been conducted. Yonekura et al. \cite{yonekura2021data} implemented conditional VAE (CVAE) in the inverse design framework to obtain the airfoil shape that satisfies specific requirements. On the other hand, Yang et al. \cite{yang2023inverse} used VAE to generate realistic pressure distribution candidates needed to perform inverse design. 

However, these studies mainly focused on solving inverse design problems rather than airfoil parameterization. To train their generative model, they used the aerodynamic coefficients \cite{yilmaz2020conditional, achour2020development, yonekura2021data} or pressure distribution \cite{yang2023inverse} with the CFD simulations, which are not always available considering the large size of the airfoil database to be represented. In addition, since the aerodynamic performance is highly dependent on the specified flow conditions such as Mach number, Reynolds number, and angle of attack, it is difficult to generalize the learned latent space.

While the aforementioned studies have primarily used generative models for inverse design tasks, there is another body of research that specifically targets the application of airfoil parameterization \cite{chen2020airfoil, du2020b, wang2023airfoil, kou2023aeroacoustic}. They considered the airfoil parameterization as a dimensionality reduction problem to derive low-dimensional representation of high-dimensional airfoil coordinates. However, an important problem witnessed in the previous studies \cite{yilmaz2020conditional, achour2020development, chen2020airfoil} is that the airfoil parameterization methods using GANs often lack feasibility: the generated airfoil geometries are self-intersecting and lack smoothness. This problem arises from the extensive flexibility inherent in the neural network, which directly affects the surface coordinates of an airfoil. It is reported that the VAE-based model mitigates these problems \cite{yang2023inverse, yonekura2021data, wang2023airfoil}), but relying on this characteristic alone may over-smooth the airfoil geometries, leading to poor flexibility. Moreover, the treatment of self-intersecting airfoil stills remained an open question.

A simple remedy for the smoothness problem is to post-process the geometry with a smoothing filter such as Savitzky–Golay \cite{savitzky1964smoothing, yilmaz2020conditional}. However, since the application of the smoothing filter is not taken into account in the training process, it can be detrimental to the objectives of the model, as reported in ref. \cite{wada2023physics}. To address this challenge, W. Chen et al. \cite{chen2020airfoil} modified the structure of GAN: Instead of directly predicting the airfoil coordinate itself, the modified GAN model first predicts the parameters of the B\'ezier curve, such as control points, and then uses these parameters to reconstruct the airfoil curve. This approach, named B\'ezierGAN, has successfully integrated the B\'ezier curve representation into the GAN architecture. Consequently, B\'ezierGAN can generate smooth airfoil shapes while retaining both the parsimony and the flexibility of deep learning-based parameterization methods. Through extensive comparative studies, they quantitatively demonstrate that B\'ezierGAN not only produces more realistic smooth airfoil geometries but also converges to optimal solutions more effectively than other techniques such as SVD and PARSEC. Building on this innovation, Lin et al. \cite{lin2022cst} took a similar approach by applying CST methods. In particular, Du et al. \cite{du2022airfoil, du2020b} further improved the B\'ezierGAN by replacing the B\'ezier layer with a B-spline layer, thus addressing the notable limitations of the B\'ezier curve in terms of local controllability.

Through their dedicated efforts, deep learning-based parameterization methods have achieved semi-feasibility while maintaining parsimony and flexibility, i.e. semi-feasibility in that the problem of self-intersecting airfoils is still unsolved. However, the limitation of these methods still lies in their lack of intuitiveness, since the neural network acts as a black-box, making the underlying processes uninterpretable. Here, interpretability means that users have quantitative and independent control over the physical features of the airfoils, which facilitates the knowledge transfer between research groups and simplifies the imposition of physical constraints. In fact, the previous studies have found that the learned latent variables of their model have interpretable features in qualitative manner due to the disentanglement nature inherent in GAN and VAE \cite{chen2020airfoil, wang2023airfoil, du2022airfoil, du2020b}. However, it remained only qualitative results, limiting their practical applicability to design optimization.

Recently, several efforts have been made to improve the transparency and interpretability of neural network models \cite{higgins2016beta, burgess2018understanding, rolinek2019variational, burda2015importance, sonderby2016ladder, takeishi2021physics}. The core of these studies is the improved understanding of the regularization effects on the latent space. It is worth noting that the diagonal Gaussian prior has interesting implications for latent space regularization. Since the diagonal covariance in Gaussian distribution implies the independence between random variables, the latent space strictly following this prior distribution aligns with the generative factors without any external information \cite{rolinek2019variational}. In addition, carefully controlled regularization suppresses irrelevant latent dimensions, i.e., the number of latent dimensions is automatically reduced to optimally represent the original dataset, leading to the minimal representation \cite{burda2015importance, sonderby2016ladder, dai2019diagnosing}. To exploit these desirable properties, $\beta$-VAE is can be employed, which uses $\beta$ as a hyperparameter to weight between the regularization loss and the reconstruction loss, thereby controlling the adherence of the latent space to the Gaussian prior.

In the field of aerodynamics, numerous researchers have applied $\beta$-VAE for the flowfield reconstruction \cite{eivazi2022towards, li2022physically, kang2022physics, wang2023physics}. For example, Eivazi et al. \cite{eivazi2022towards} utilized the method to extract the physically interpretable modes from the turbulent flowfield. Kang et al. \cite{kang2022physics} applied the $\beta$-VAE to transonic flowfield dataset and investigated the effect of $\beta$ on the extraction of interpretable latent variables. In addition, Li et al. \cite{li2022physically} added a physical loss term to directly inform the physical features on the latent dimension, so that the latent variables are closely aligned with the user-defined physical features.

By exploiting these ``physics-aware'' variational autoencoders, this paper develops a physically intuitive airfoil parameterization while preserving other desired properties of deep learning-based generative models (cf. While variants of $\beta$-VAEs are referred to using various terms such as ``physically-interpretable'' \cite{lu2018improved}, ``physics-aware'' \cite{kang2022physics}, ``physics-assisted'' \cite{wang2023physics}, etc., this study simply adopts the term ``physics-aware'' for convenience). To achieve these goals, two strategies were employed. First, the generation of thickness and camber distributions is explicitly separated using an independent encoding/decoding architecture. This not only allows the independent control of thickness and camber distributions, but also completely prevents the generation of self-intersecting airfoils. Second, the learned latent dimensions are directly aligned with the geometric features of the airfoil, such as maximum thickness/camber, leading edge radius, and trailing edge angle. This alignment is extensively validated using both qualitative and quantitative measures. In particular, a novel regularization technique using latent sampling is proposed to improve the generalizability of the physically aligned latent variables. Furthermore, this study benchmarks the architecture of BsplineGAN \cite{du2020b, du2022airfoil}, which wraps the decoder with a B-spline layer to ensure the smoothness of the generated airfoils. With these modifications, the proposed airfoil parameterization method, namely Airfoil Generator, satisfies all the desirable properties: flexibility, parsimony, feasibility, and intuitiveness. In order to validate the performance of the Airfoil Generator, an extensive comparative study with existing methods has also been conducted using various benchmark problems, including inverse fitting, unconstrained/constrained airfoil shape optimization.

The remainder of this paper is organized as follows. Section \ref{sec:background} briefly describes the existing airfoil parameterization methods, laying the foundation for our subsequent discussions. Section \ref{sec:methodology} presents the main architectures of our proposed method. Section \ref{sec:results} presents an extensive comparative study that carefully evaluates our method against the four critical properties of flexibility, feasibility, parsimony and intuitiveness. Finally, section \ref{sec:conclusion} summarises our findings and outlines future research.

\section{Background Knowledge}\label{sec:background}

To ensure the self-containment of this paper, this section provides a brief overview of several key airfoil parameterization techniques: PARSEC \cite{sobieczky1999parametric}, CST \cite{kulfan2008universal} B\'ezier/B-splines \cite{piegl1996nurbs}, SVD \cite{toal2010geometric}, and VAE \cite{kingma2013auto}. Readers familiar with these methods can skip to the following sections. For a more in-depth exploration, please consult the references for each method.

\subsection{Conventional Polynomial-based Parameterizations}
\subsubsection{Parameterized Section (PARSEC)}
Developed by Sobieczky \cite{sobieczky1999parametric}, PARSEC's design variables correlate closely with the physical characteristics of the airfoil, such as leading edge radius and the locations of upper/lower crest. In PARSEC, both the upper and lower surfaces are represented by a sixth order polynomial:

\begin{equation}\
\label{eq:parsec}
\begin{aligned}
\boldsymbol{Y_u}(\boldsymbol{X}) = \sum_{i=1}^{6} a_{u,i}\boldsymbol{X}^{i-0.5}\\
\boldsymbol{Y_l}(\boldsymbol{X}) = \sum_{i=1}^{6} a_{l,i}\boldsymbol{X}^{i-0.5}
\end{aligned}
\end{equation}

where $a_i$ represents the coefficient of each order of polynomial. These coefficients are determined by solving twelve equations, corresponding to twelve airfoil-related design variables. Similar techniques are also performed in B\'ezier-PARSEC \cite{derksen2010bezier} and IGP \cite{lu2018improved}. However, unlike the PARSEC, which separates the airfoil with upper and lower surfaces, both methods represent the airfoil with the thickness and camber distributions. These methods are significant in that their design variables are directly related to the physical features of the airfoil. However, due to the nature of solving a linear system, their behavior is often ill-conditioned, resulting in infeasible geometries \cite{wu2003comparisons}. In addition, the degree of the polynomial is typically fixed, limiting the flexibility of the parameterization method \cite{masters2017geometric}.

\subsubsection{Class/Shape-Function Transformation (CST)}
Kulfan \cite{kulfan2008universal} introduced the CST method, a framework for representing airfoil shapes using basis functions. The surfaces are defined by a class function scaled by a shape function.

\begin{equation}
\label{eq:cst}
\begin{aligned}
\boldsymbol{Y}(\boldsymbol{X}) &= \underbrace{\boldsymbol{X}^{N_1}(1-\boldsymbol{X})^{N_2}}_{C(\boldsymbol{X}):\, Class\,  function} \cdot \underbrace{\sum_{i=0}^{n} a_iB_{i,n}(\boldsymbol{X})}_{S(\boldsymbol{X}):\,  Shape\, function}
\end{aligned}
\end{equation}

where $C(\boldsymbol{X})$ is the class function and $S(\boldsymbol{X})$ is the shape function. For airfoils, the class function assumes $N_1=0.5$ and $N_2=1$ to obtain a rounded leading edge and a sharp trailing edge, respectively. The shape function is represented by the Bernstein polynomial of degree $n$, $B_{i,n}$, which is defined as follows:

\begin{equation}\
\label{eq:Bernstein_poly}
\begin{aligned}
B_{i,n}(\boldsymbol{X}) = \binom{n}{i} \boldsymbol{X}^i (1-\boldsymbol{X})^{n-i}
\end{aligned}
\end{equation}

\subsubsection{B\'ezier curve}
The B\'ezier curve is a popular method of representing a curve using control points. Given the scalar parameter $u \in [0,1]$, the B\'ezier curve is defined by the following equation:

\begin{equation}\
\label{eq:bezier}
\begin{aligned}
B(u) = \sum_{i=0}^{n} B_{i,n}(u) \cdot \boldsymbol{P}_i
\end{aligned}
\end{equation}

where $\boldsymbol{P}_i$ are control points, $n$ is the degree of the B\'ezier curve, and $B_{i,n}$ is the basis of the Bernstein polynomials of degree $n$. However, as implied by the weighted sum form of Eq. \ref{eq:bezier}, changing a single control point affects the entire curve, potentially limiting flexibility.

\subsubsection{B-splines}

B-splines are a generalization of the B\'ezier curve. While the B\'ezier curve uses the Bernstein polynomial as the basis function, B-splines compute the basis function using the Cox-de Boor algorithm. Through this modification, B-splines gain the local control on the curve. Given a knot vector $ \boldsymbol{U} = \{ u_0, u_1, ..., u_m \} $, the B-spline basis functions $ N_{i,p}(u) $ of degree $p$ are defined recursively as:

For $p = 0$:\

\begin{equation}\
\label{eq:bspline_1}
\begin{aligned}
N_{i,0}(u) = \begin{cases}
1, & \text{if } u_i \leq u < u_{i+1} \\
0, & \text{otherwise}
\end{cases}
\end{aligned}
\end{equation}

For $ p > 0 $:\

\begin{equation}
\label{eq:bspline_2}
\begin{aligned}
N_{i,p}(u) = \frac{u - u_i}{u_{i+p} - u_i} N_{i,p-1}(u) + \frac{u_{i+p+1} - u}{u_{i+p+1} - u_{i+1}} N_{i+1,p-1}(u)
\end{aligned}
\end{equation}

Finally, B-splines are represented by the following equation:

\begin{equation}\
\label{eq:bsplin_3}
\begin{aligned}
B(u) = \sum_{i=0}^{n} N_{i,p}(u) \cdot \boldsymbol{P}_i
\end{aligned}
\end{equation}

When the degree of B-spline basis $p$ is equivalent to $n$ (where $n+1$ is the number of control points), and the knot vector is clamped and uniform with no internal knots, the B-spline basis functions reduce to Bernstein polynomials.

\subsection{Data-driven Parameterizations}
\subsubsection{Singular Value Decomposition (SVD)}
SVD \cite{toal2010geometric} is one of the most popular dimensionality reduction techniques that can extract the most dominant modes from the given dataset. Therefore, when applied to airfoil parameterization, it can effectively extract parsimonious design variables to represent high-dimensional airfoil coordinates. SVD utilizes eigenvector/eigenvalue decomposition, which is represented by the following equation:

\begin{equation}
\label{eq:svd_eigendcompose}
\begin{aligned}
\boldsymbol{M} = \boldsymbol{U} \cdot \boldsymbol{\Sigma} \cdot \boldsymbol{\Phi}
\end{aligned}
\end{equation}

where $\boldsymbol{M}$ represents the covariance matrix of the differences between airfoil y-coordinates, $\boldsymbol{\Sigma}$ and $\boldsymbol{\Phi}$ represent the eigenvalue and eigenvector, respectively. Since the extracted modes are orthogonal and ordered, the original airfoils can be reconstructed as a linear combination of these modes $\varphi_i$:

\begin{equation}
\label{eq:svd}
\begin{aligned}
\boldsymbol{Y} \approx \sum_{i=1}^{m}{a_i\boldsymbol{\varphi_i}} + \bar{\boldsymbol{Y}} \quad (m<<d)
\end{aligned}
\end{equation}

where $\boldsymbol{\bar{Y}}$ represents the extracted mean to zero-center the dataset, and $d$ and $m$ is the original and the reduced number of dimensions, respectively.

\subsubsection{Variational Autoencoder (VAE)}\label{sec:background:vae}
\begin{figure}[htb!]
	\centering
	
	\begin{subfigure}[b]{0.475\linewidth}
        \centering
        \includegraphics[width=\linewidth]{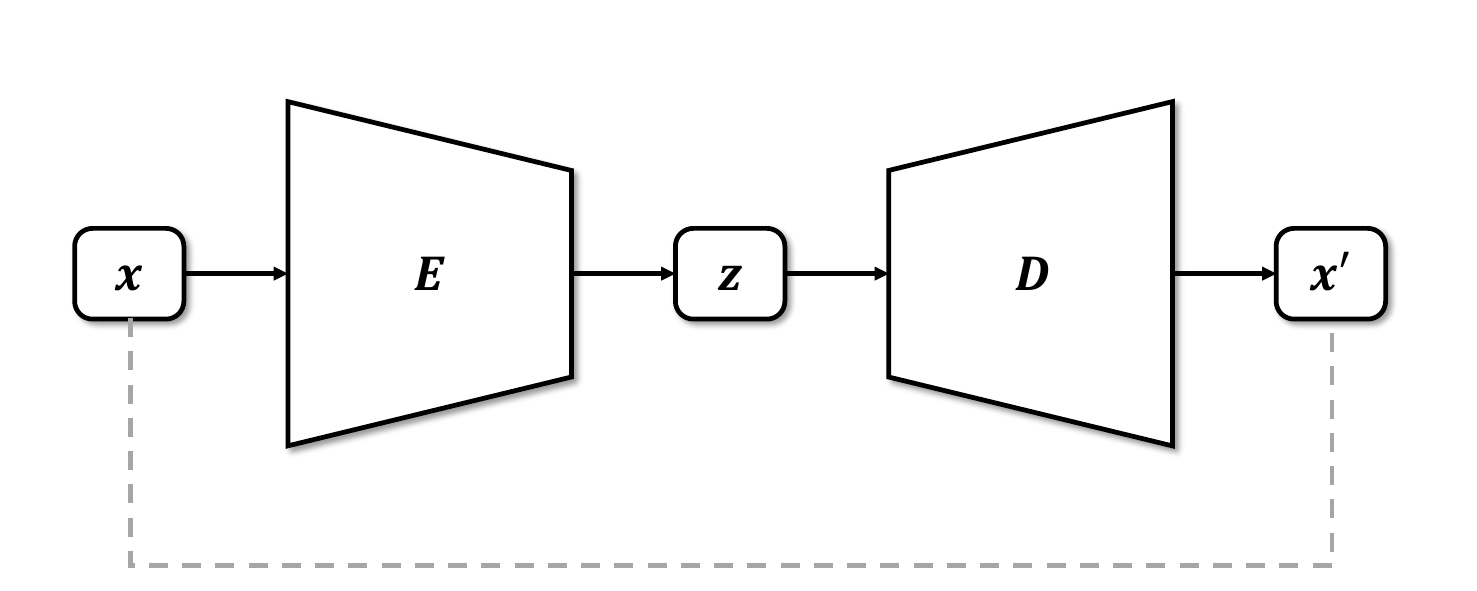}
        \caption{Autoencoder (AE)}
        \label{fig:model_archi_AE}
    \end{subfigure}
    \hfill 
    \begin{subfigure}[b]{0.475\linewidth}
        \centering
        \includegraphics[width=\linewidth]{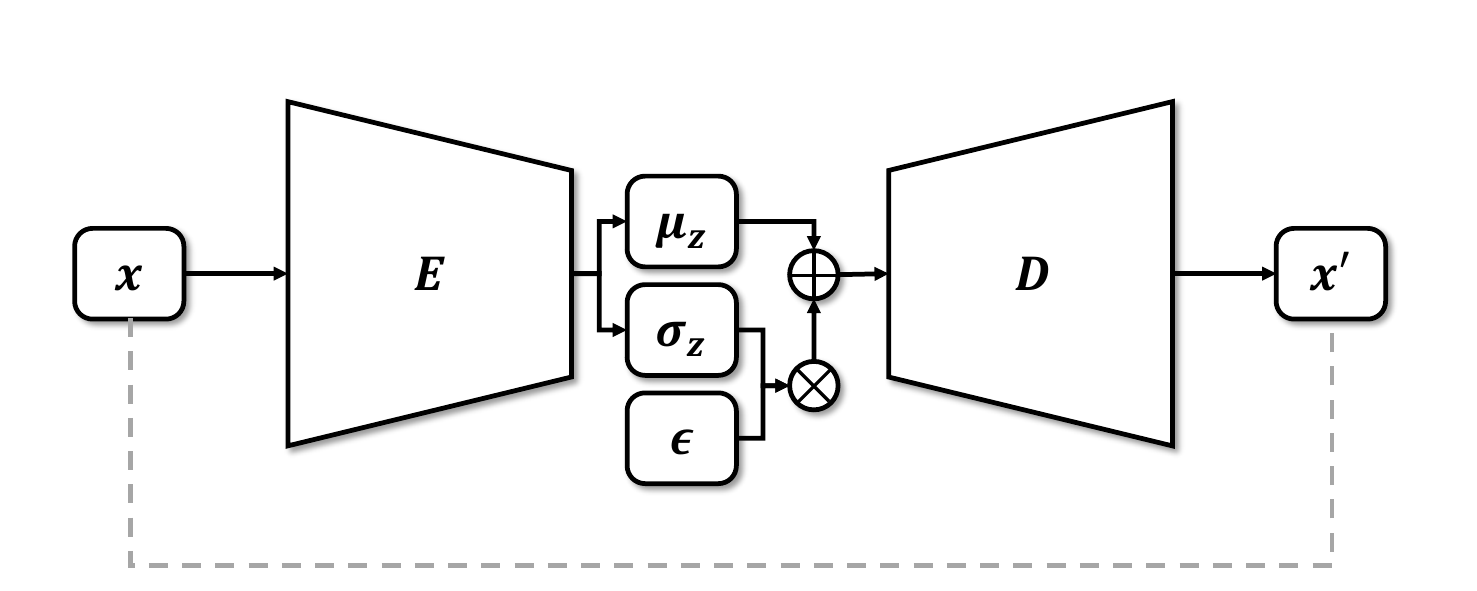}
        \caption{Variational autoencoder} 
        \label{fig:model_archi_VAE}
    \end{subfigure}

    \vspace{1em} 
    
    \begin{subfigure}[b]{0.475\linewidth}
        \centering
        \includegraphics[width=\linewidth]{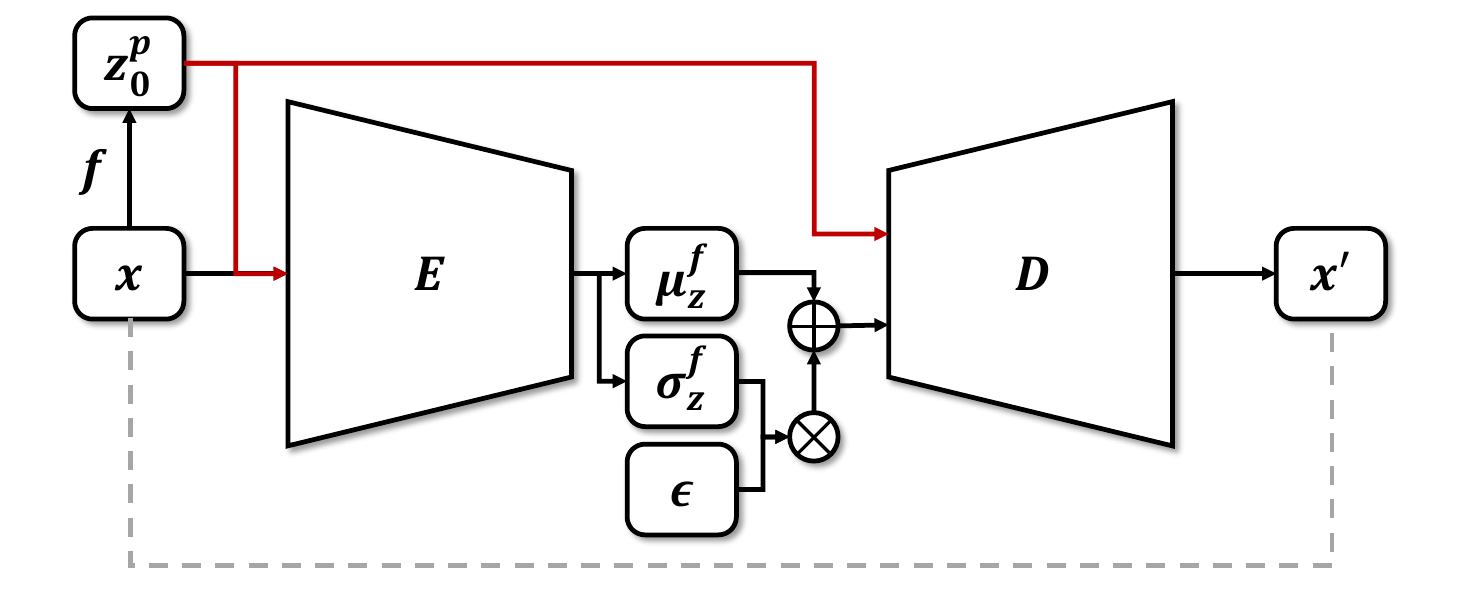}
        \caption{Conditional variational autoencoder}
        \label{fig:model_archi_CVAE}
    \end{subfigure}
    \hfill 
    \begin{subfigure}[b]{0.475\linewidth}
        \centering
        \includegraphics[width=\linewidth]{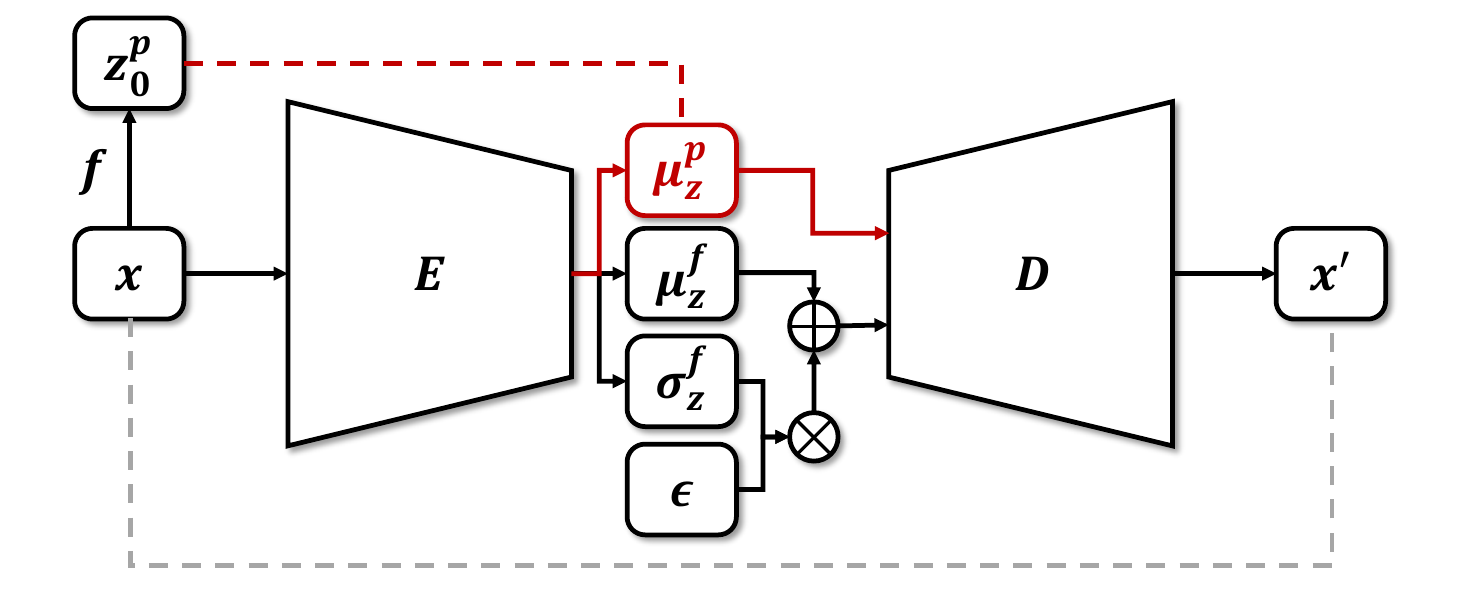}
        \caption{Phyically-interpretable variational autoencoder \cite{li2022physically}}
        \label{fig:model_archi_PIVAE}
    \end{subfigure}
    
    \caption{Comparison of the variants of variational autoencoder}
    \label{fig:model_archi_VAEs_tot}
\end{figure}

Recently, there have been several attempts to leverage VAE for both airfoil parameterization and inverse design \cite{kou2023aeroacoustic, yonekura2021data, li2022physically}. In fact, variational autoencoders share structural similarities with autoencoders (AEs), which is the neural network model for extracting latent features from high-dimensional data. As shown in Figs. \ref{fig:model_archi_AE} and \ref{fig:model_archi_VAE}, both architectures use a bottleneck structure, consisting an encoder $E$ that compresses a data sample $x$ and a decoder $D$ that reconstructs it to resemble original sample $x'$. However, since AE does not impose any constraints on the latent space, it is not suitable as a generative model. In contrast, VAE imposes constraints on the latent space to conform to prior distributions $p(z)$, typically a standard normal distribution. This requirement allows VAE to be more effective as generative model. 

In VAEs, the encoder and decoder are represented by the probabilistic distributions $q(z|x)=\mathcal{N}(\mu_z, \sigma_z * I)$ and $p(x|z)=\mathcal{N}(\mu_x, \sigma_x * I)$, respectively, which are both assumed as Gaussian distributions parameterized by neural networks. The training objective of VAE is to maximize the log-likelihood of the observed data, denoted by $\log(p(x))$. However, direct computation of log-likelihood is often intractable because it requires integration over latent variables. To address this problem, variational inference is used to optimize the evidence lower bound (ELBO), which is the lower bound of the log-likelihood:

\begin{equation}
\label{eq:vae_elbo}
\begin{aligned}
\text{ELBO} = \underbrace{\mathbb{E}_{\boldsymbol{z} \sim q(\boldsymbol{z}|\boldsymbol{x})} \left[ \log p(\boldsymbol{x}|\boldsymbol{z}) \right]}_{\text{Reconstruction loss}} - \underbrace{D_{\text{KL}} \left( q(\boldsymbol{z}|\boldsymbol{x}) || p(\boldsymbol{z}) \right)}_{\text{Regularization loss}}
\end{aligned}
\end{equation}

The ELBO consists of two components: the reconstruction loss and the regularization loss. Since the VAE assumes the encoder output is Gaussian, the reconstruction loss is reduced to the mean squared error (MSE) between the original samples and their reconstructions, denoted as $\text{MSE}_\text{recon} = \frac{1}{N_s}\Vert\boldsymbol{x}-\boldsymbol{x}'\Vert^2_2$, where $N_s$ is the number of samples. The regularization loss, represented by Kullback-Leibler divergence (KLD), encourages the encoded latent distribution $q(z|x)$ to approximate prior distribution $p(z)$. Given the importance of balancing these two losses, a weighting parameter $\beta$ is introduced to adjust the trade-off between reconstruction and regularization. This variant of the VAE is commonly referred to as the $\beta$-VAE, whose loss function can be represented by the following equation:

\begin{equation}
\label{eq:beta_vae_loss}
\begin{aligned}
\mathcal{L}_{\beta-\text{VAE}} = \frac{1}{N_s}\Vert\boldsymbol{x}-\boldsymbol{x}'\Vert^2_2 + \beta \cdot D_{\text{KL}} \left( q(\boldsymbol{z}|\boldsymbol{x}) || p(\boldsymbol{z}) \right)
\end{aligned}
\end{equation}

It is worth noting that the regularization loss term has a significant impact on the VAE, in terms of the interpretability of the model. It encourages the latent space to conform to the Gaussian distribution with diagonal covariance, which facilitates the disentanglement of the latent variables \cite{rolinek2019variational}. It also forces the deactivation of latent dimensions that do not contribute to the reconstruction, resulting in a minimal representation \cite{burda2015importance, sonderby2016ladder, dai2019diagnosing}. 

While VAE works without any external label information that can be computed from the data, there is another variant of VAE that can take the physical label into account, called conditional VAE (CVAE). As shown in Fig. \ref{fig:model_archi_CVAE}, the CVAE utilizes label data $z_0^p$ obtained from training samples as additional input to both the encoder and decoder. CVAE also maximizes the ELBO as in VAE, but in a conditional form:

\begin{equation}
\label{eq:cvae_elbo}
\begin{aligned}
\text{ELBO} = \mathbb{E}_{z \sim q(\boldsymbol{z}|\boldsymbol{x}, z_0^p)} \left[ \log p(\boldsymbol{x}|\boldsymbol{z}, z_0^p) \right] - D_{\text{KL}} \left( q(\boldsymbol{z}|\boldsymbol{x}, z_0^p) || p(\boldsymbol{z}) \right)
\end{aligned}
\end{equation}

When the training is finished, the decoder is able to generate samples conditioned on specific labels, expressed as $\boldsymbol{x'} \sim p(\boldsymbol{x}|\boldsymbol{z}, z_0^p)$.

Recently, Li et al. \cite{li2022physically} developed a new variant of VAE, called physically-interpretable VAE (PIVAE). In this model, as shown in Fig. \ref{fig:model_archi_PIVAE}, latent variables are divided into two groups: physical and free (denoted by superscript $p$ and $f$, respectively). The physical latent variables are designed to be aligned with known physical labels. They proved that the alignment can be achieved by implementing a physical MSE in the VAE loss function. This term penalizes discrepancies between the physical latent variables and the physical labels. The PIVAE loss function is defined as follows:

\begin{equation}
\label{eq:pivae_loss}
\begin{aligned}
\mathcal{L}_{\text{PIVAE}} = \frac{1}{N_s}\Vert\boldsymbol{x}-\boldsymbol{x}'\Vert^2_2 + D_{\text{KL}} \left( q(\boldsymbol{z}^f|\boldsymbol{x}) || p(\boldsymbol{z}^f) \right) + \Vert\boldsymbol{\mu_z}^p-\boldsymbol{z}^p_0\Vert^2_2
\end{aligned}
\end{equation}

The difference between the CAVE and PIVAE architectures is that CVAE takes the physical labels as input to both the encoder and the decoder. In contrast, PIVAE first predicts the physical labels using the encoder and then takes them as input for the decoder. Li et al. confirmed that PIVAE outperforms CVAE in terms of generating physically constrained design samples, especially when the training label deviates from the uniform distribution \cite{li2022physically}. Inspired by this PIVAE structure, our research aimed to develop an airfoil parameterization method with improved feasibility and intuitiveness, which will be presented in the following sections.

\section{Airfoil Parameterization via Physics-aware Variational Autoencoder}\label{sec:methodology}
\subsection{Model Architecture}

While existing deep learning-based parameterizations offer the advantage of high flexibility and fewer design variables, they often fall short in terms of feasibility and intuitiveness. To address these limitations, this study has made several extensions to the standard VAE framework. This section delves into the modified structure by dividing it into the three primary components: encoding, decoding, and bottleneck.

\begin{figure*}[htb!]
	\centering
		\includegraphics[width=1.\textwidth]{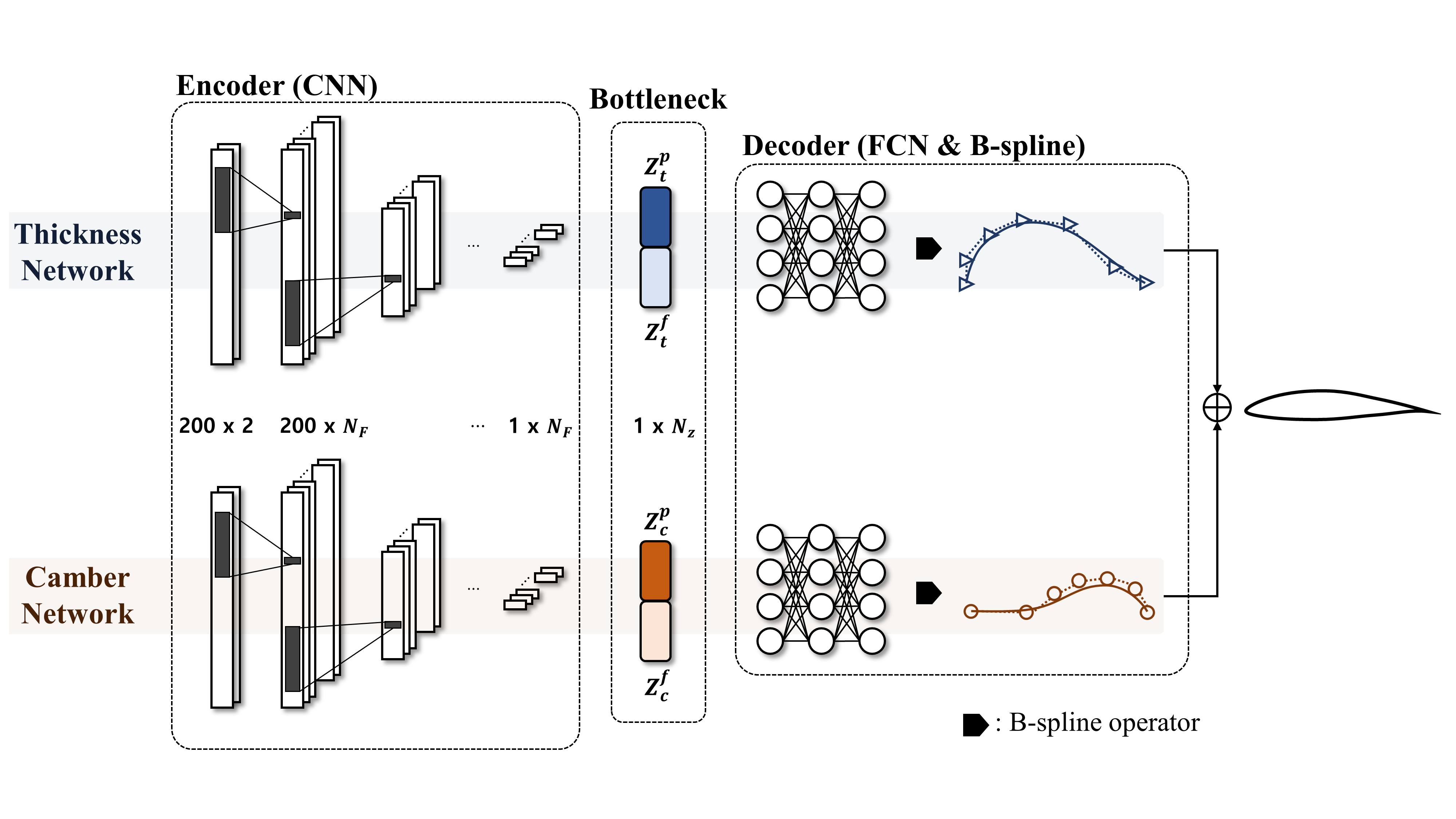}
	\caption{Structures of Airfoil Generator.}
	\label{fig:model_architecture}
\end{figure*}

\subsubsection{Encoding Part}
The role of the encoder is to extract dominant features from the original high-dimensional data, which in this study are the x and y coordinate vectors of the airfoil. In general, an autoencoder has a bottleneck structure. This design ensures that the dimensionality of the input data is progressively reduced, allowing the high-dimensional data to be substantially compressed into a low-dimensional latent space. As a result, this compressed representation can serve as design variables with the potential to represent airfoils.

As shown in Fig. \ref{fig:model_architecture}, the Airfoil Generator consists of two encoders: one for the thickness distribution and another for the camber distribution. This "two-way" structure, which separates the generation of the thickness and camber distributions, is adopted for two main reasons. First, the dominant features of the thickness and camber distributions can be extracted independently. Due to their separate latent spaces, users can control both distributions independently, each of which has a different effect on aerodynamic performance. Second, by ensuring that the thickness distribution remains positive, the generation of self-intersecting airfoils can be completely avoided. This consideration is further discussed in the following section.

For the neural network architecture of the encoder, we used a one-dimensional convolutional neural network (CNN). Since the neurons in the CNN are connected to only a small region of the subsequent layer \cite{o2015introduction}, the number of learnable weights is much smaller than in a fully connected layer, thus avoiding overfitting. This locality is also suitable for airfoil coordinates where there are strong correlations exist between neighboring grid points. To further improve the performance of the encoder, several key hyperparameters were carefully chosen: the size of the filter/kernel, the dimension of the latent space, and the activation functions, which were tuned using a grid search algorithm.

\subsubsection{Decoding Part}
The decoder serves as a central component in the airfoil parameterization, reconstructing the original airfoil geometry from parsimonious latent variables. In general, the structure of the decoder is symmetrical to that of the encoder. However, when applied to airfoil reconstruction, the neural network that directly controlling the coordinate is susceptible to generate noisy and non-smooth airfoil geometry due to the high flexibility of the neural network, as already confirmed in ref \cite{chen2020airfoil}. 

To mitigate this problem, Chen et al. \cite{chen2020airfoil} integrated the B\'ezier curve representation into the decoder. Instead of directly reconstructing the airfoil coordinate, they first reconstruct the parameters of the B\'ezier curve, such as control points, parameteric location, and the weights of control points, and then reconstruct the original airfoil shape using the B\'ezier representation system. Through this approach, their neural network model can be guaranteed to generate the smooth geometry. Du et al. \cite{du2020b, du2022airfoil} further improved the model by replacing the B\'ezier curve with a B-spline, as the B-spline has more local control on the curve than the B\'ezier curve, allowing the higher flexibility.

However, even though the B\'ezier and B-spline layers enforce smooth geometry, self-intersecting airfoils are often observed, especially near the trailing edge where the thickness is very small. Both previous models attempt to solve this problem by adding a penalty term in the loss function. However, modifying the loss function will not completely eliminate the generation of self-intersecting airfoils and will compromise other loss terms such as airfoil reconstruction error.

Therefore, instead of adding penalty term in the loss function, this study introduced simple but straightforward way to generate non-intersecting airfoils; explicit decomposition of the thickness and camber distributions. The thickness distribution $\boldsymbol{Y}_t(\boldsymbol{X})$ and the camber distribution $\boldsymbol{Y}_m(\boldsymbol{X})$ are defined as the following equations:

\begin{equation}\
\label{eq:tc_decompose}
\begin{aligned}
\boldsymbol{Y}_t(\boldsymbol{X}) = \boldsymbol{Y}_u(\boldsymbol{X})-\boldsymbol{Y}_l(\boldsymbol{X}) \\
\boldsymbol{Y}_c(\boldsymbol{X}) = \frac{1}{2}[\boldsymbol{Y}_u(\boldsymbol{X})+\boldsymbol{Y}_l(\boldsymbol{X})]
\end{aligned}
\end{equation}

\begin{figure*}[htb!]
	\centering
\includegraphics[width=0.6\textwidth]{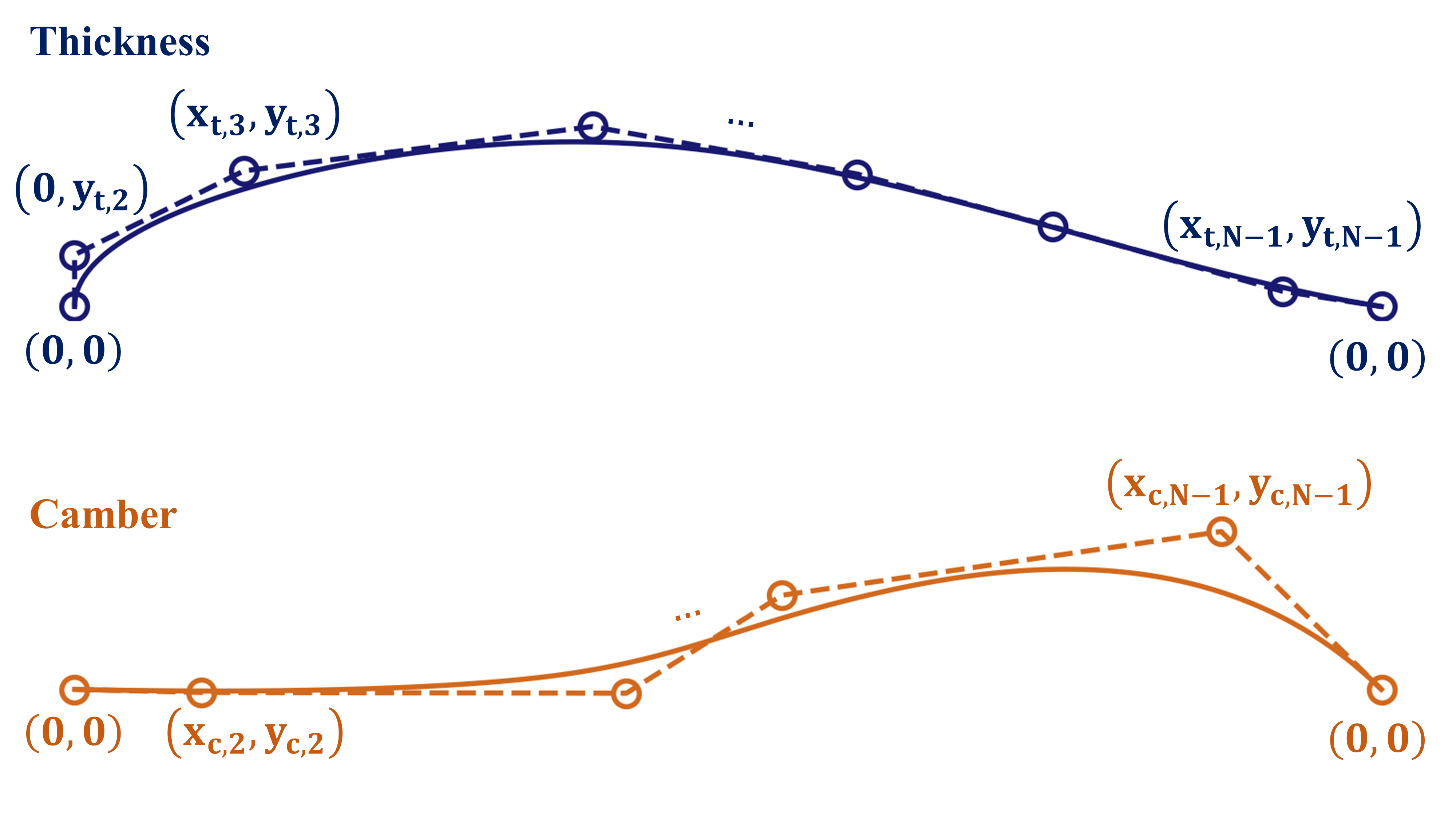}
	\caption{Regularization of B-spline control points.}
	\label{fig:Bspline_ctr}
\end{figure*}

By properly regularizing the control points of the b-splines to generate a positive thickness distribution, this approach can completely prevent the generation of self-intersecting airfoils. The details of control point regularization are illustrated in Fig. \ref{fig:Bspline_ctr}. Suppose that there are sets of control points $\boldsymbol{P_t} = \{(x_{t,1}, y_{t,1}), (x_{t,2}, y_{t,2}), ..., (x_{t,N}, y_{t,N}))\} $ and $\boldsymbol{P_c} = \{(x_{c,1}, y_{c,1}), (x_{c,2}, y_{c,2}), ..., (x_{c,N}, y_{c,N}))\} $, where the subscripts $t$ and $c$ indicate the thickness and camber, respectively. We have assumed that the leading edge and trailing edge of airfoils are fixed at $(0,0)$ and $(0,1)$, respectively, and that the x-coordinates of each control point are non-decreasing. This results in $0=x_{1}\leq x_{2}\leq...\leq x_{N}=1$ and $y_{1}=y_{N}=0$. Additionally, for the control points of the thickness distribution, the x-coordinate of the second control point was also set to zero to avoid a sharp leading edge $(x_{t,2}=0)$ and all of the y-coordinates were set to non-negative for non-intersecting airfoils $(y_{t,2}, y_{t,3}, ..., y_{t,N-1} \geq 0)$. Consequently, the number of free variables for thickness and camber are $2N_t-3$ and $2N_c-2$, respectively. In fact, the number of control points can significantly affect to the representative capability. Through trial and error, it was confirmed that the representative capability reaches sufficient accuracy when the number of control points for both thickness and camber distribution is 12 with the degree of 3.

The neural network in the decoder correlates the latent space with b-spline control points. A fully connected network was used to satisfy the constraints described above. For the non-decreasing x-coordinate constraint, the neural network predicts the x-axis-wise distance between each point rather than directly predicting the x-coordinate itself. For the positive y-coordinate constraint of the thickness distribution, absolute operation is applied prior to the b-spline operation.

\subsubsection{Bottleneck Part}

As discussed in Section \ref{sec:background:vae}, VAE can be used as a generative model by assuming a prior distribution of the latent space. The common prior distribution is multivariate Gaussian distribution with diagonal covariance, and it is implemented as a KL-divergence loss function. It is reported that the regularization of latent space  facilitates the disentangled and parsimonious latent representation \cite{kang2022physics}. Therefore, VAE-based Airfoil Generator can acheive independent and parsimonious airfoil parameterization. However, geometric features of the airfoil, such as maximum camber \( m_{\text{max}} \), trailing edge direction \( \gamma_{\text{TE}} \), maximum thickness \( t_{\text{max}} \), and leading edge radius \( r_{\text{LE}} \), is neither always independent each other nor Gaussian distributed. Consequently, the learned latent space does not necessarily align with these physical features. This presents an obstacle to fully exploiting the intuitive design variables in the design optimization process.

\begin{figure*}[htb!]
	\centering
\includegraphics[width=0.7\textwidth]{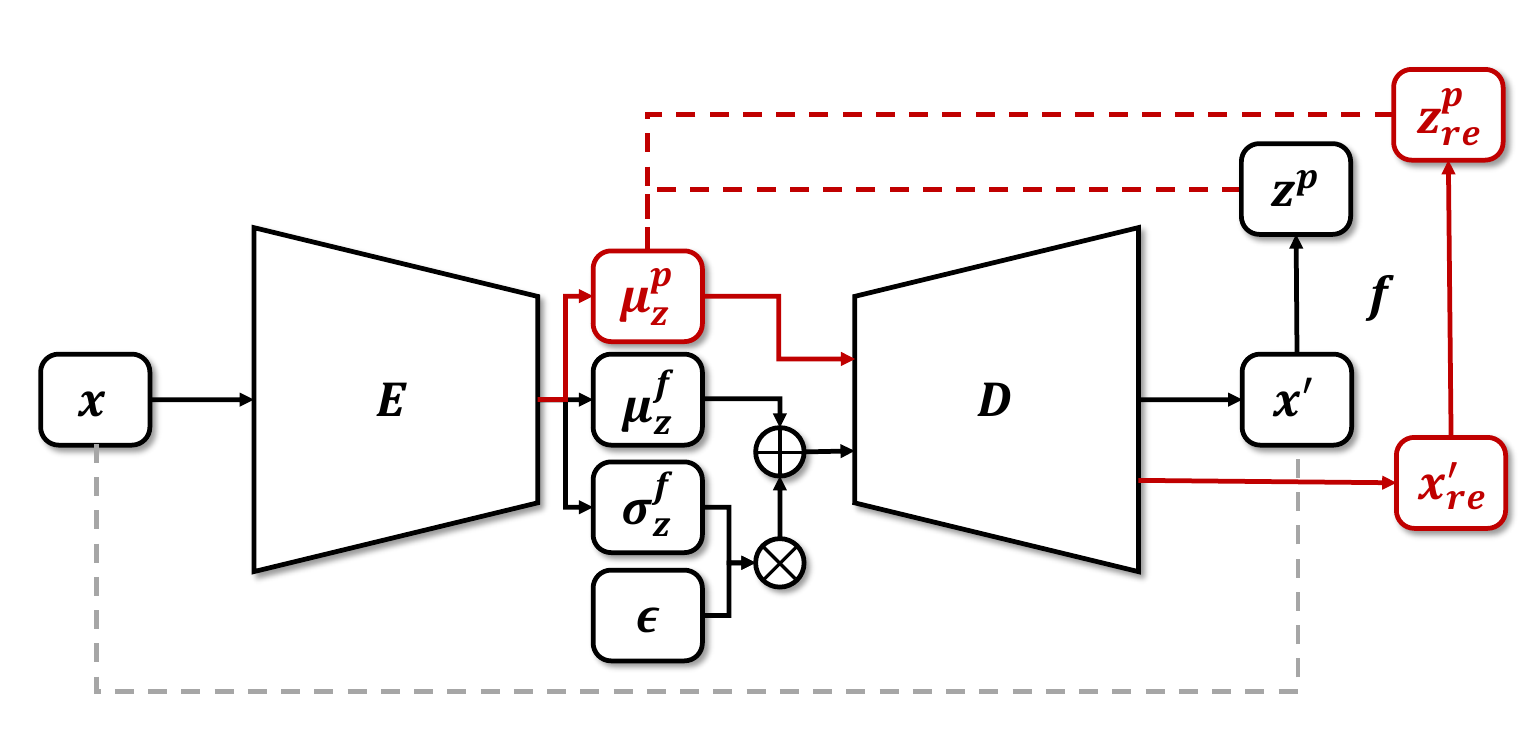}
	\caption{Implementation of $\text{MSE}_\text{phys}$ in the proposed VAE model.}
	\label{fig:model_archi_present}
\end{figure*}

To address these issues, we first benchmark the architecture of PIVAE \cite{li2022physically}, with a subtle modification to improve the model, as shown in Fig. \ref{fig:model_archi_present}. As in PIVAE, we first split the latent dimensions into "physical" and "free" latent dimensions, represented as $\mu_{z}^{p}$ and $\mu_{z}^{f}$, and then force the physical latent dimensions to align with the physical features. This alignment is achieved by including a mean square error term \( \text{MSE}_{\text{phys}} \) in the loss functions, the theoretical background of which can be found in ref. \cite{li2022physically}.

\begin{equation}\
\label{eq:physloss}
\begin{aligned}
\text{MSE}_\text{phys} = \Vert \boldsymbol{\mu_{z}}^{p}-\boldsymbol{z}^p \Vert^2_2
\end{aligned}
\end{equation}

where $z^p$ represents the user-defined physical features that can be calculated from airfoil geometries. In this study, four physical features were selected; $m_\text{max}$, $\gamma_\text{TE}$ for camber distribution, and $t_\text{max}$, $r_\text{LE}$ for thickness distribution. These geometric features are properly normalized to range from 0 to 1  before training, denoted as $\hat{m}_\text{max}$, $\hat{\gamma}_\text{TE}$, $\hat{t}_\text{max}$, and $\hat{r}_\text{LE}$. It is worth noting that the physical features are calculated from the generated sample $x'$ rather than the training sample $x$, which differs from the original PIVAE architecture shown in Fig. \ref{fig:model_archi_PIVAE}. There are two reasons for this choice: First, since the core part of the Airfoil Generator is a decoder part that generates the airfoil geometry from the latent variables, it is more consistent to align the latent variables with the generated airfoil rather than the training airfoil. Second, since the Airfoil Generator uses B-splines for curve representation, direct access to the derivative of the thickness/camber distribution is available, which is very useful when calculating physical properties such as leading edge radius.

\begin{figure*}[htb!]
	\centering
\includegraphics[width=0.45\textwidth]{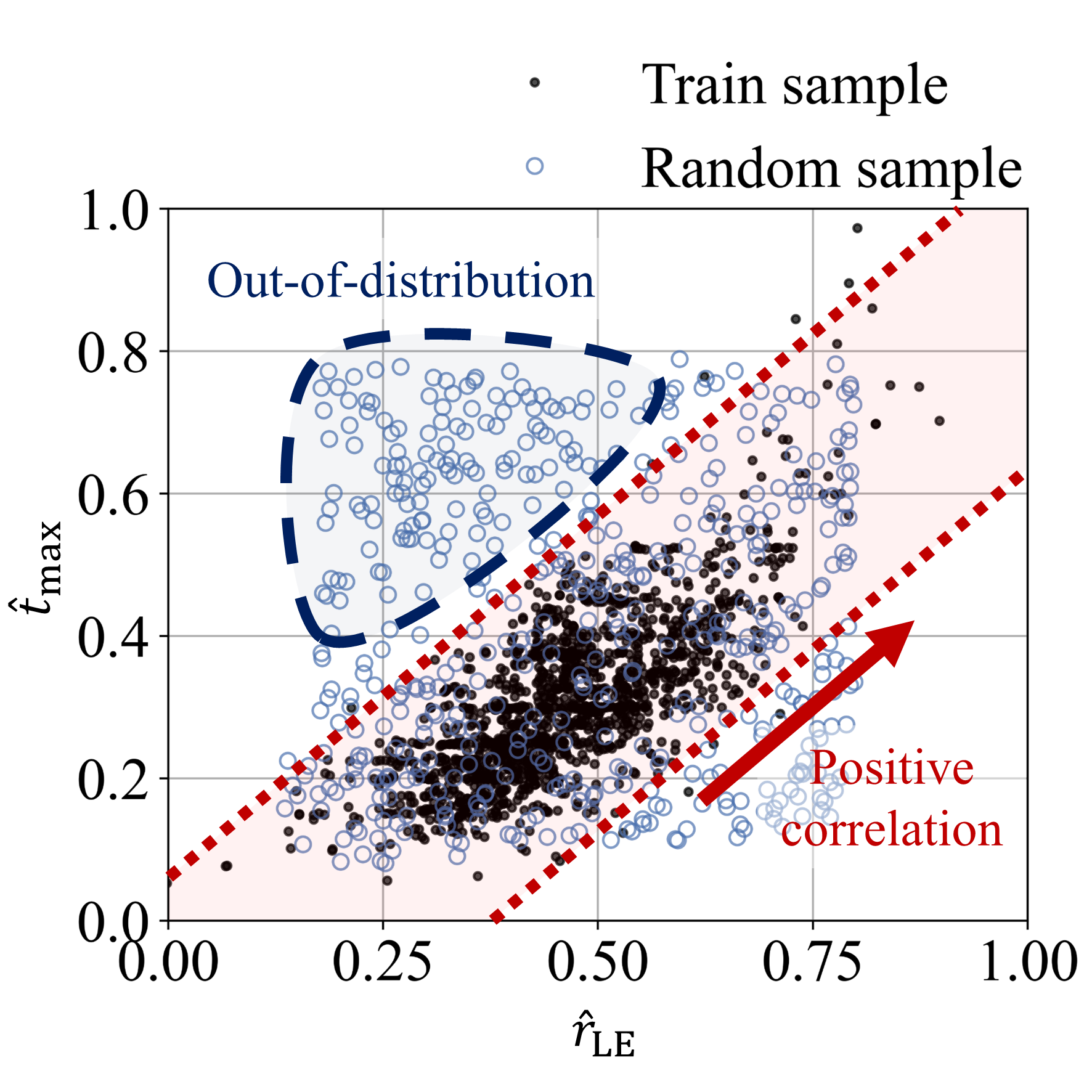}
	\caption{Visualization of out-of-distribution in the $\hat{r}_\text{LE}$-$\hat{t}_\text{max}$ plane.}
	\label{fig:Visualization_OOD}
\end{figure*}

However, when applying this physical loss, we have found that there is a risk of overfitting, where only the physical latent variables of the training samples are fitted to the geometric features. This arises from the fact that the physical features of real-world data are likely to be correlated, whereas the learned latent dimensions are expected to be independent. This mismatch leads to out-of-distribution (OOD) regions within the latent space - regions that the training data does not cover. For example, Fig. \ref{fig:Visualization_OOD} is the visualization of OOD region in $\hat{r}_{le}$-$\hat{t}_{max}$ plane. The black dots represent the UIUC airfoil sample used for model training, and the blue circles represent the generated sample with the random combination of latent variables. From the figure, we can see that there is a positive correlation between the training samples. However, when we consider the latent dimensions as independent of each other, there are OOD regions that the UIUC airfoil samples cannot cover. Although these regions are a source of novel design candidates, their detachment from the training sample space often results in the generation of design candidates that are of sub-optimal quality.

To prevent the overfitting, this study proposes a novel regularization technique, called ``latent sampling regularization''. This strategy consists in taking into account the random samples $x'_\text{re}$ when calculating $\text{MSE}_\text{phys}$ (cf. The subscript $\text{re}$ denotes resampling). Note that the advantage of the generative model is that the realistic samples can be generated by latent sampling. Considering that $\text{MSE}_\text{phys}$ is the term related only to the latent variables and the corresponding geometry, airfoils from latent sampling can also be used as training samples to calculate this loss, $\text{MSE}_\text{phys}^\text{random}$. Therefore, during the training process, the physical latent loss is calculated on both UIUC and latent sampled airfoils. In summary, the loss function of the proposed model is as follows:

\begin{equation}
\label{eq:ag_loss}
\begin{aligned}
\mathcal{L}_{\text{AG}} = \frac{1}{N_s}\Vert\boldsymbol{x}-\boldsymbol{x}'\Vert^2_2 + \beta \cdot D_{\text{KL}} \left( q(\boldsymbol{z}^f|\boldsymbol{x}) || p(\boldsymbol{z}^f) \right) + \lambda \cdot \left( \Vert\boldsymbol{\mu_z}^p-\boldsymbol{z}^p\Vert^2_2 + \Vert\boldsymbol{\mu_{z_{re}}}^p-\boldsymbol{z}^p_{re}\Vert^2_2 \right)
\end{aligned}
\end{equation}

where $\beta$ and $\lambda$ are weighting parameters to balance the loss terms. In this study, $\beta$ is considered as a hyperparameter and is determined using a grid search algorithm. Meanwhile, the value of $\lambda$ is set to $10^{-5}$ by trial and error to ensure that it is small enough not to significantly affect other loss terms. Latent sampling is performed at each epoch, and the boundary of the bounding box is determined based on the 0.3\% to 99.7\% of the latent variables of the training samples.

\subsection{Data Preparation and Normalization}
To train the Airfoil Generator, the UIUC airfoil database \cite{selig1996uiuc} are used, which is the historical collection of airfoils consisting of over 1500 geometry samples. The airfoils in this database cover a wide range of characteristics: the maximum camber, \( m_{\text{max}} \), ranges from 0 to 0.15, while the maximum thickness, \( t_{\text{max}} \), varies from 0.02 to 0.66.

Since the data format of the UIUC database varies from sample to sample (e.g. the number of data points), appropriate data formatting and normalization are required before training the Airfoil Generator. In this study, the raw data was first interpolated with a spline. Next, the leading and training edges were aligned to (0,0) and (0,1), and the points were again distributed by interpolation so that the upper and lower surfaces of the airfoil have the same x-coordinate. The x-coordinate is defined by the following equation:

\begin{equation}\
\label{eq:cos_distribution}
\begin{aligned}
x_i = (1-\cos(\frac{\pi}{100}i))/2,\quad i= 0, 1,..., 100
\end{aligned}
\end{equation}

Finally, poorly defined airfoils were excluded by visual inspection. A total 1539 samples are remained. Among them, 1499 samples were used for training, and the rest of them were used for validation/testing.

\subsection{Training Details}
An appropriate selection of hyperparameters is crucial for efficient compression/reconstruction of airfoils. Therefore, hyperparameter tuning based on a grid search was performed to determine suitable structures. For hyperparameters, the size of filter $(N_\text{filter})$,  kernel $(N_\text{kernel})$, the dimension of latent space $(N_\text{latent})$, activation function, and regularization parameter $(\beta)$ were considered. However, full-factorial sampling of these hyperparameters is computationally too intensive. Therefore, grid search was performed sequentially: first, $\beta$ was fixed to zero, and $N_\text{filter}$, $N_\text{kernel}$, $N_\text{latent}$, activation function were varied. After finding optimal settings for these four hyperparameters, the grid search was performed again on $\beta$. 

The grid points for hyperparameter tuning are detailed in the Table. \ref{tab:hyperparameters}. For the optimization settings, the Adam algorithm has been selected as the optimizer. The initial learning rate is set to $10^{-2}$ and it decays at a rate of 0.7 every 2500 epochs for a total of 25,000 epochs. The whole training samples were used as a batch, because it showed more stable behavior during the training process.

\begin{table}[htb!]
\centering
\setlength\tabcolsep{20pt} 
\caption{Grid points for hyperparameter tuning.}
\label{tab:hyperparameters}
\begin{tabular}{lc}
\hline \hline
\textbf{Hyperparameter} & \textbf{Values} \\
\hline
\( N_{\text{filter}} \) & 32, 64, 128 \\
\( N_{\text{kernel}} \) & 3, 7 \\
\( N_{\text{latent}} \) & 6, 12 \\
Activation Function & GELU, ReLU, LeakyReLU \\
\( \beta \) & \( 10^{-9} \) to \( 10^{-7} \) (11 points in log scale) \\
\hline \hline
\end{tabular}
\end{table}

\begin{figure*}[htb!]
	\centering
		\includegraphics[width=1.\textwidth]{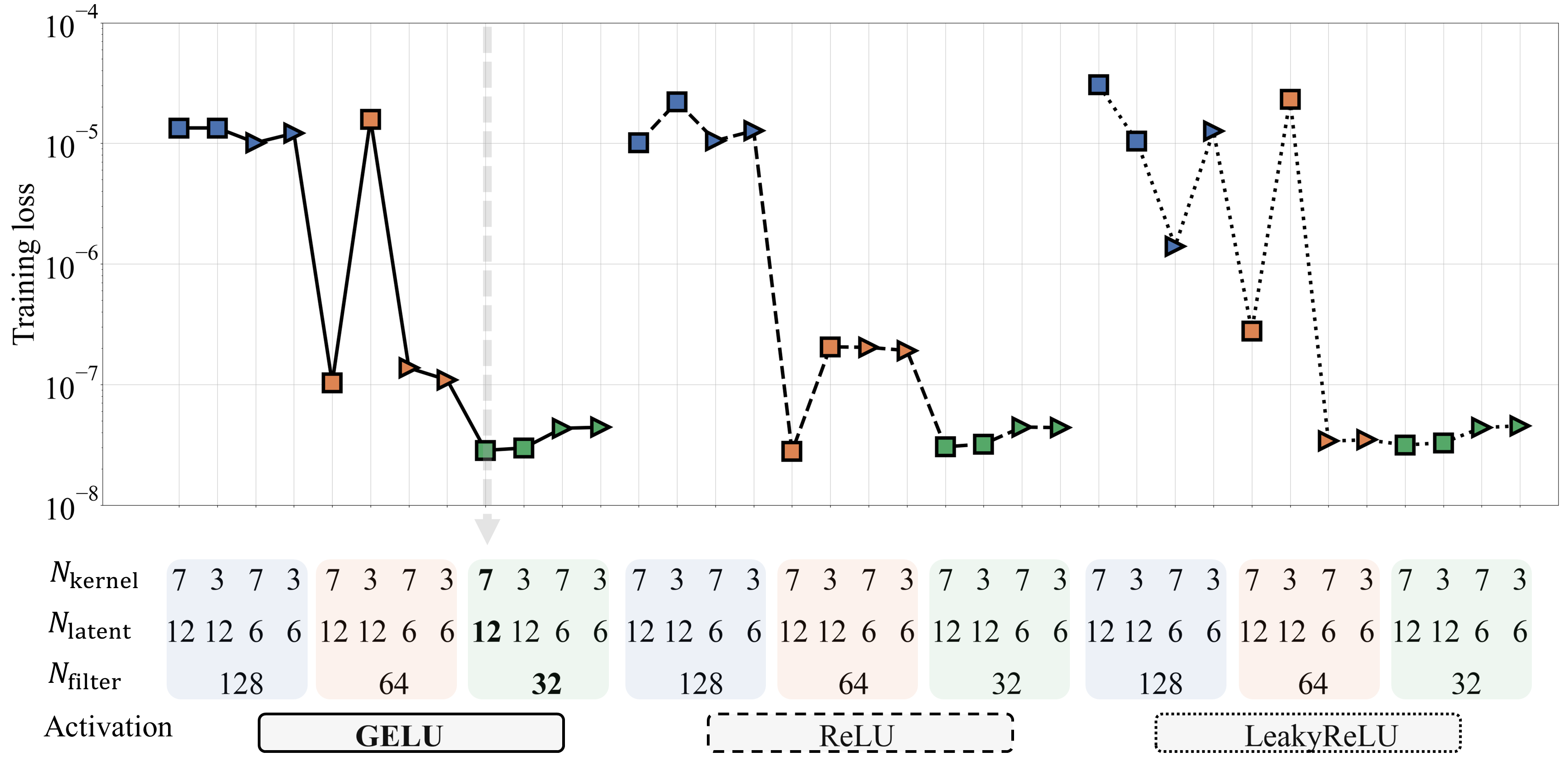}
	\caption{Hyperparameter tuning results of encoder.}
	\label{fig:hyperparameter_tuning}
\end{figure*}

\begin{figure*}[htb!]
	\centering
		\includegraphics[width=0.7\textwidth]{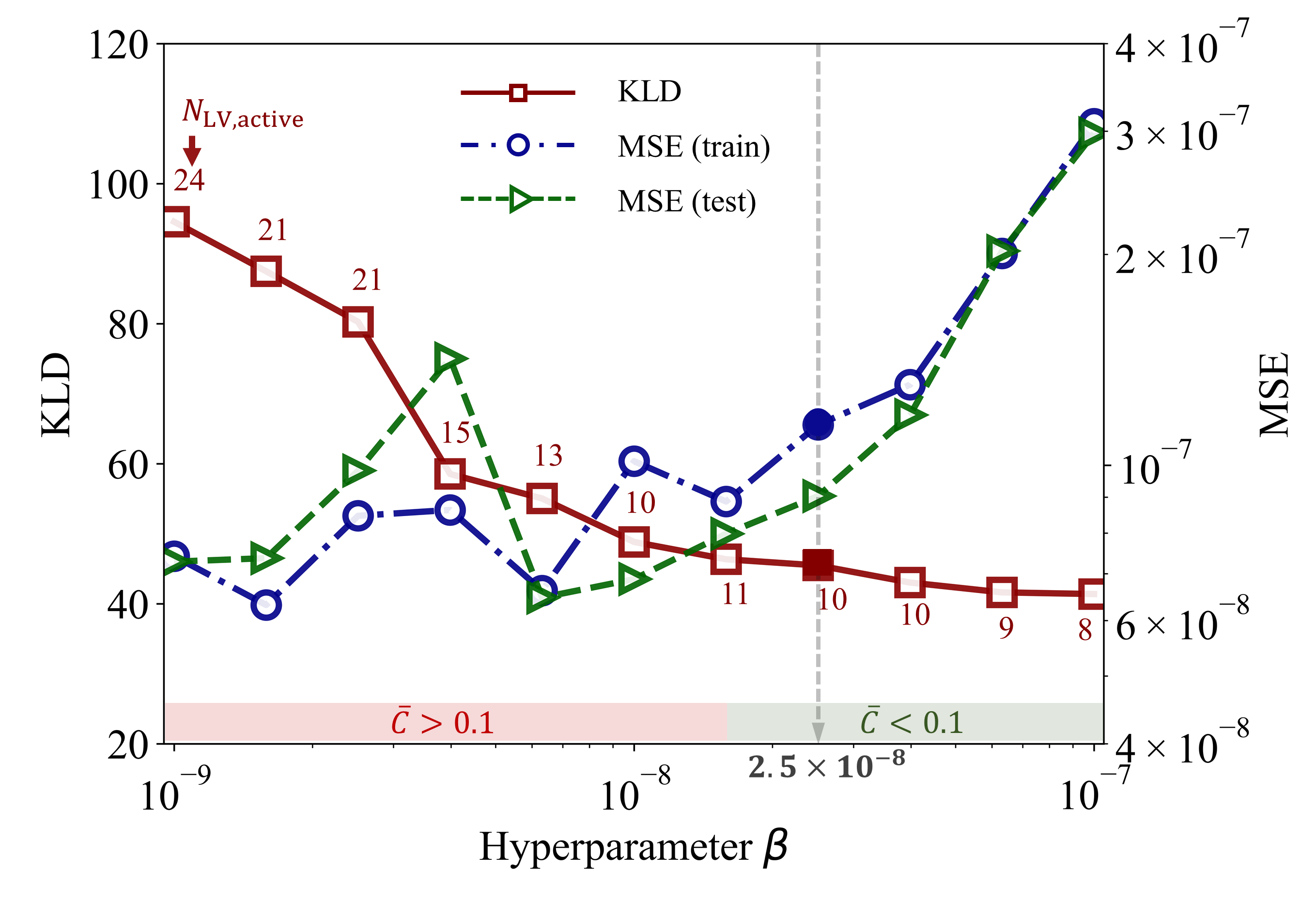}
	\caption{Mean-squared-error (MSE) and Kullback-Leibler divergence (KLD) with respect to the hyperparameter $\beta$.}
	\label{fig:beta_tuning}
\end{figure*}

The hyperparameter tuning results are shown in Figs. \ref{fig:hyperparameter_tuning} and \ref{fig:beta_tuning}. Figure \ref{fig:hyperparameter_tuning} shows the first hyperparameter tuning results, where the x-axis represents the index of each model and the y-axis represents the training loss to be minimized. The results show that $N_\text{kernel}$, $N_\text{latent}$, and the type of activation function have no significant impact on the training loss. However, $N_\text{filter}$ showed a significant effect. Specifically, a smaller filter size resulted in better performance in terms of training loss. It is interpreted that large filter size has negative effect on convergence due to the large number of weight parameters to be optimized. Therefore, in this study, the hyperparameter setting of $N_\text{kernel}=7, N_\text{filter}=32, N_\text{latent}=12$ and GELU was selected, which showed the lowest training loss.

Figure \ref{fig:beta_tuning} shows the results of hyperparameter tuning on $\beta$. The x-axis represents $\beta$ and the y-axis represent the KLD and MSE, respectively. It can be observed that there is trade-off between KLD and MSE; the smaller the KLD, the larger the MSE. It is worth noting that in $\beta$-VAE, several latent dimensions are automatically deactivated by the pressure to minimize the KLD \cite{kang2022physics}. In this study, the influence of each latent dimension is measured by the average standard deviation of the y-coordinate of the output airfoil when only one latent dimension is swept while the other latent dimensions are held constant. The number of active latent variables $(N_\text{LV, active})$ is defined as the number of latent variables of which mean standard deviation is larger than $10^{-5}$, and the results are written next to the red solid line in Fig. \ref{fig:beta_tuning}. It can be observed that $(N_\text{LV, active})$ gradually decreases with KLD (note that both physical and free latent variables are counted together in the number).

\begin{figure*}[htb!]
	\centering
		\includegraphics[width=1.0\textwidth]{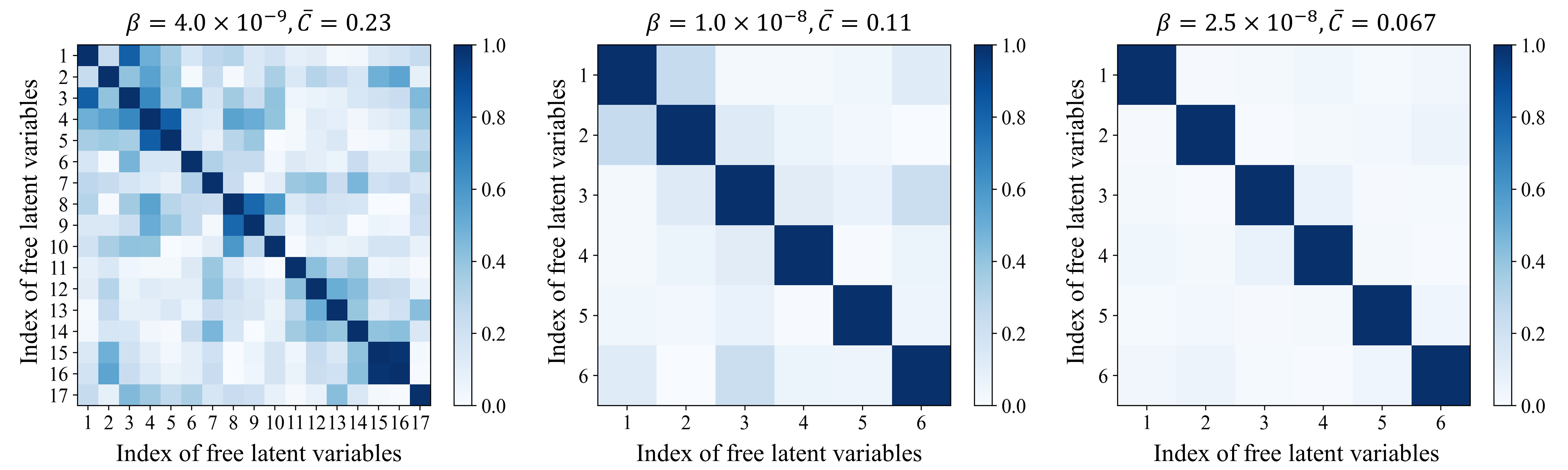}
	\caption{Pearson correlation matrix with respect to the hyperparameter $\beta$.}
	\label{fig:Pearson}
\end{figure*}

To determine the optimal value of $\beta$, we monitored the Pearson correlation matrix of free latent dimensions, which measures the independence of latent dimensions. The mean value of correlation coefficient $\bar{C}$ was calculated as a quantitative criterion, which is defined as follows:

\begin{equation}\
\label{eq:mean_correlation_matrix}
\begin{aligned}
\bar{C} = \frac{\sum_{i=2}^{N_\text{DV}} \sum_{j=1}^{i-1} c_{ij}}{N_\text{DV}(N_\text{DV}-1)/2}
\end{aligned}
\end{equation}

where $c_{i,j}$ represents correlation between $i^{th}$ and $j^{th}$ free latent variables, and $N_\text{DV}$ represents the number of design variables. Figure \ref{fig:Pearson} is the visualization of the Pearson correlation matrix, where each axis represents the index of the latent dimensions and the heatmap represents the absolute value of the Pearson coefficients. It is confirmed that the increment of $\beta$ results in decrement of correlation coefficient, which means more independent latent dimensions. It is interesting that even the number of active free latent variables are the same at $\beta=1.0 \times 10^{-8}$ and $2.5 \times 10^{-8}$, $\bar{C}$ is much lower at $\beta=2.5 \times 10^{-8}$ than $\beta=1.0 \times 10^{-8}$ (note that, at $\beta=1.0 \times 10^{-8}$, some pairs of the latent dimensions are not yet independent). Therefore, the case of $\beta=2.5 \times 10^{-8}$ was chosen as a reference model. We also considered the model with $\beta=1.0 \times 10^{-7}$ for comparison, to investigate the relationship between parsimony and flexibility. The detailed results are presented in the following section.

\subsection{Physical latent variable}
\begin{figure}[htb!]
	\centering
	\begin{subfigure}[b]{1.\linewidth}
        \centering
        \includegraphics[width=\linewidth]{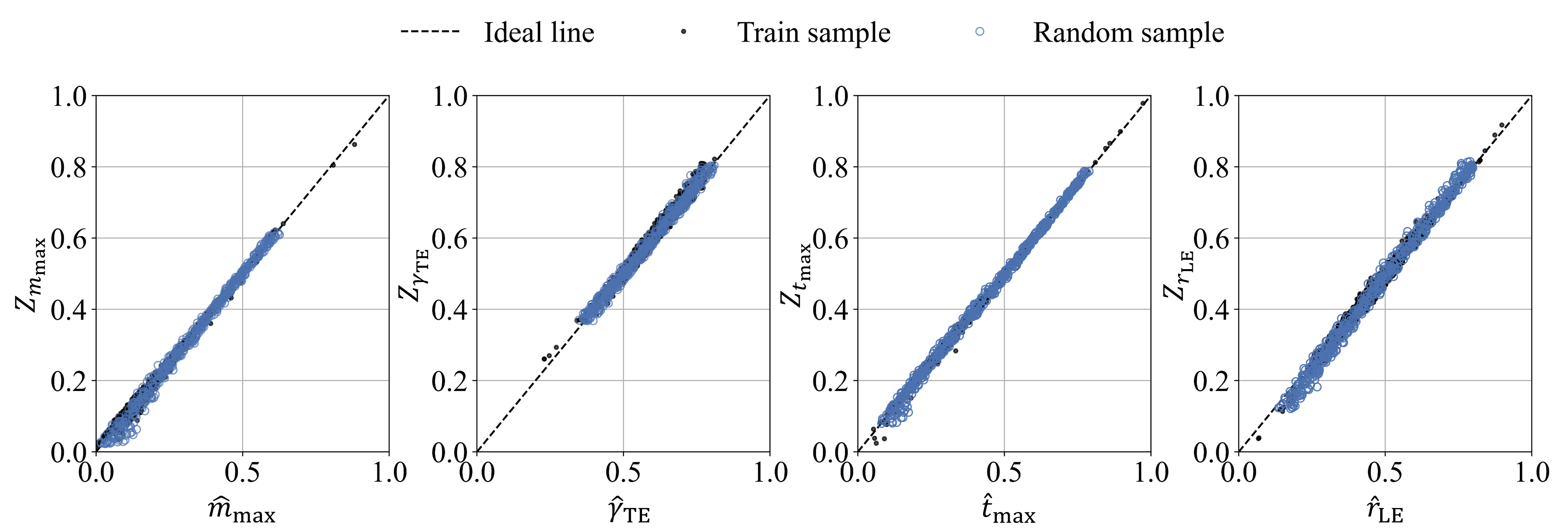}
        \caption{With latent sampling regularization}
        \label{fig:CV_phys_w_MSE_rand}
    \end{subfigure}
    \vspace{1em} 

    \begin{subfigure}[b]{1.\linewidth}
        \centering
        \includegraphics[width=\linewidth]{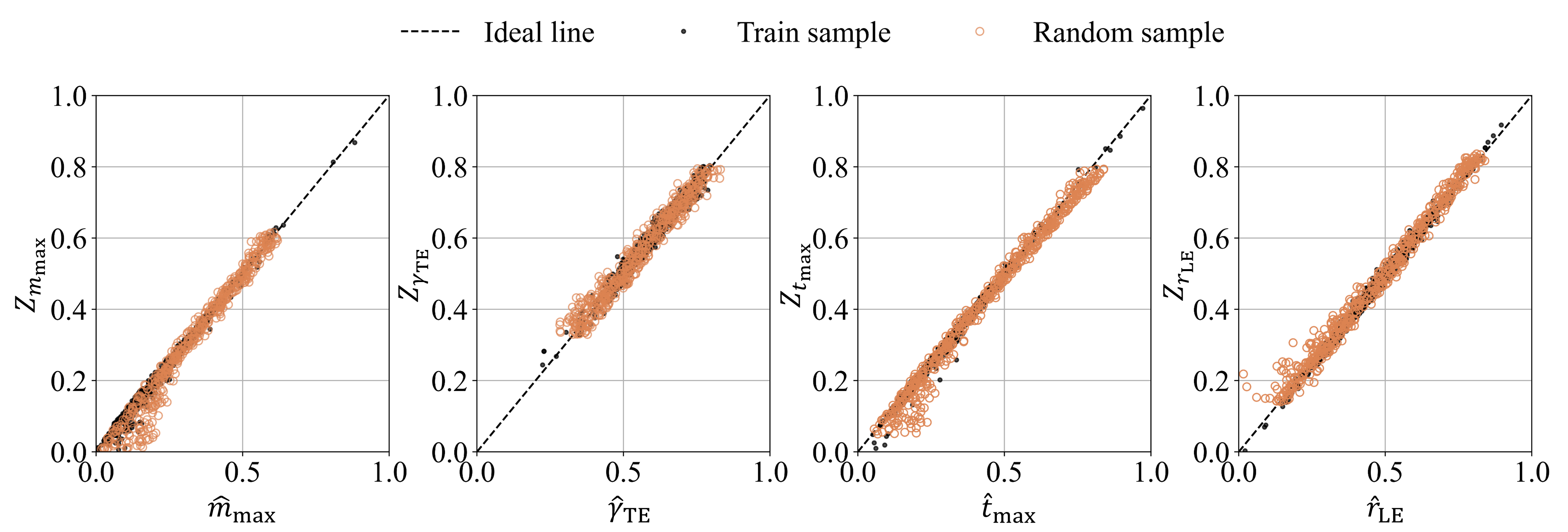}
        \caption{Without latent sampling regularization}
        \label{fig:CV_phys_wo_MSE_rand}
    \end{subfigure}
    \vspace{1em} 
    \caption{Cross-validation of physical latent variables}
    \label{fig:CV_phys}
\end{figure}

To improve the intuitiveness of the Airfoil Generator, we ensure a direct mapping between the latent variables and the physical properties using the physics loss function $\text{MSE}_\text{phys}$. A cross-validation is performed to validate the alignment between the physical latent variables and the physical features. In Fig. \ref{fig:CV_phys}, both training cases with/without latent sampling regularization are presented to confirm the effect of the regularization. The x-axis represents the normalized physical features calculated directly from the generated airfoils and the y-axis represents the physical latent variables. In the figures, the training samples are represented by the black dot and the samples that randomly generated in the bounding box are represented by the colored round marker. 

A notable observation from Fig. \ref{fig:CV_phys_w_MSE_rand} is that both training and random samples are close to the ideal line, which clearly indicates that the physical latent variables are well aligned with the physical features of the airfoil. However, in Fig. \ref{fig:CV_phys_wo_MSE_rand}, the case without latent sample regularization, only the trained samples are close to the ideal line, while the random samples deviate far from the ideal line. This can also be quantitatively confirmed by the $\text{MSE}_\text{phys}$ error in Table. \ref{tab:latent_sampling}. Both methods, with and without latent sampling, show similar MSE values for the training data ($\text{MSE}_\text{phys}^\text{train}$). The values are 1.31E-4 and 1.19E-4 respectively, indicating that both methods perform comparably well for the training data. The most notable difference is the $\text{MSE}_\text{phys}$ for the randomly sampled data ($\text{MSE}_\text{phys}^\text{random}$), which is much lower at 2.72E-4 with the latent sample regularization, compared to 9.87E-4 without it. This suggests that the latent sampling strategy significantly improves the generalization capabilities of the model. These results demonstrate the importance of latent sample regularization for training physical latent variables.

\begin{table}[htb!]
\centering 
\caption{Comparison of $\text{MSE}_\text{phys}$ values with and without latent sampling.}\label{tab:latent_sampling}
\setlength\tabcolsep{20pt} 
\begin{tabular}{ccc} 
\hline \hline
& \multicolumn{2}{c}{\textbf{Latent Sampling}} \\
\cline{2-3}
\textbf{Metric} & \textbf{w/} & \textbf{w/o} \\ \hline
$\text{MSE}_\text{phys}^\text{train}$       & 1.31E-4     & 1.19E-4      \\
$\text{MSE}_\text{phys}^\text{random}$        & 2.72E-4     & 9.87E-4      \\  
\hline \hline
\end{tabular}
\end{table}

\section{Results and Discussion}\label{sec:results}

In this section, the Airfoil Generator is compared with other parameterization methods, such as SVD \cite{toal2010geometric}, CST \cite{kulfan2008universal}, B\'ezier \cite{piegl1996nurbs}, PARSEC \cite{sobieczky1999parametric}, and IGP\cite{lu2018improved}, in terms of parsimony, flexibility, feasibility, and intuitiveness. For parsimony and flexibility (in Sec. \ref{sec:parsimony_flexibility}), inverse fitting is conducted on target airfoils for different parameterization methods with different number of design variables. For feasibility and intuitiveness (in Sec. \ref{sec:feasibility_intuitiveness}), the properties are compared using both quantitative and qualitative measures, such as the latent traversal plot, the feasibility ratio, and the maximum correlation coefficients. Finally, in order to comprehensively compare the efficiency of each parameterization method, airfoil shape optimizations for both unconstrained and constrained problems are conducted in Sec. \ref{sec:ASO}

\subsection{Parsimony and Flexibility}\label{sec:parsimony_flexibility}
In order to evaluate the parsimony and the flexibility of the airfoil parameterization methods, inverse fitting was conducted for the given target airfoil. A total of 300 target airfoils were randomly selected from the UIUC airfoil database. The design variables that best fit the target airfoils were searched using genetic algorithm. The population size is set to 200 per generation, and the optimization is terminated when the improvement of MSE is below $10^{-9}$ during the period of 200 generations. The maximum number of generation is set to 5,000. Since the number of design variables can significantly affect the flexibility, the design variables of each parameterization method have been unified to 8 and 10.

\begin{figure*}[htb!]
	\centering
\includegraphics[width=0.6\textwidth]{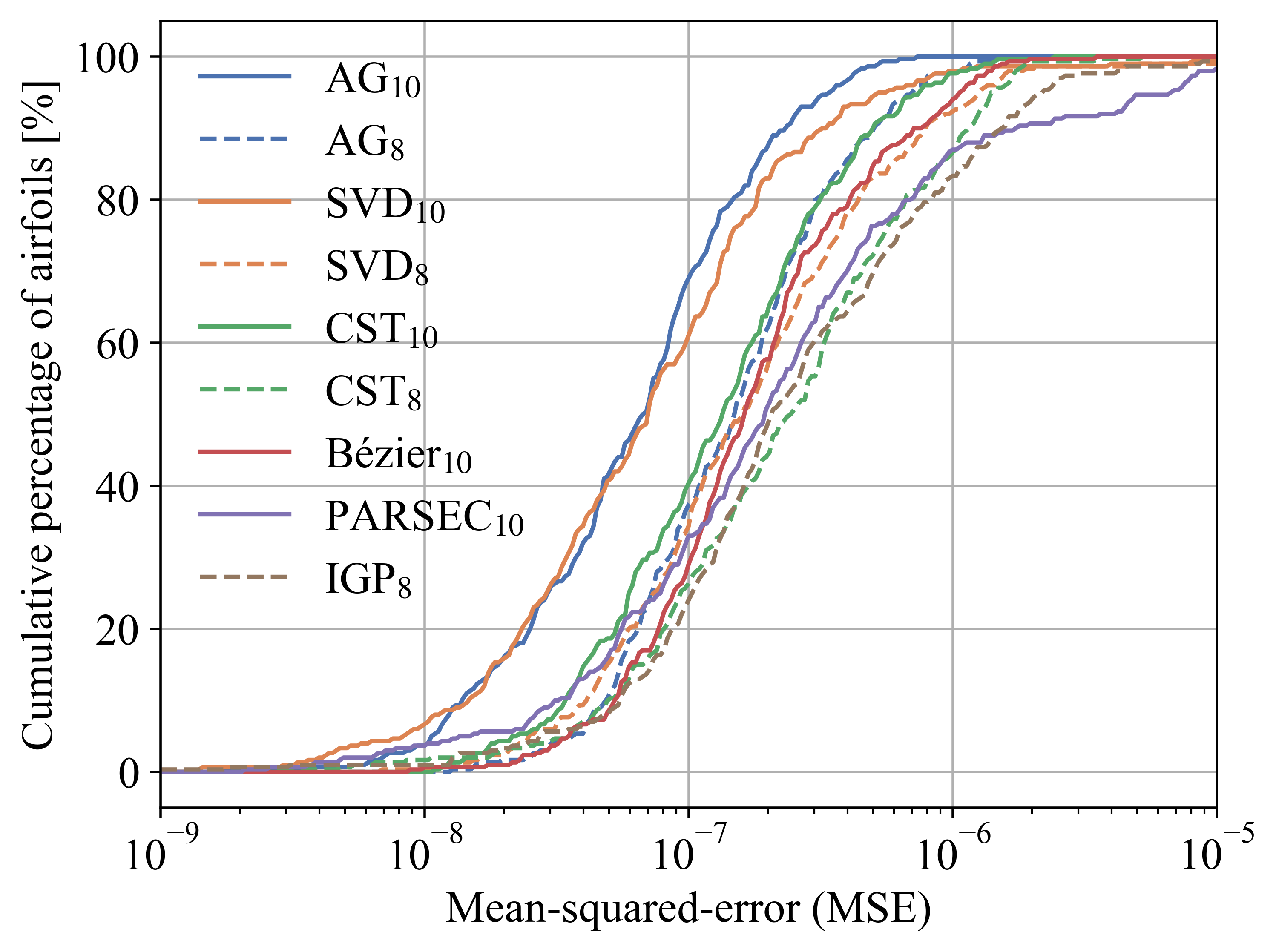}
	\caption{Cumulative distribution of mean-squared-error for inverse fitting problem.}
	\label{fig:inverse_fitting}
\end{figure*}

Figure \ref{fig:inverse_fitting} shows the results of the inverse fitting, where the x-axis represents MSE and the y-axis represents the cumulative percentage of airfoil samples. For example, if the cumulative percentage is $60\%$ at an MSE of $10^{-7}$, it means that the parameterization method can represent $60\%$ of the target airfoil database within the error bound of $10^{-7}$. Each subscription on the parameterization method label indicates the number of design variables. The results confirm that Airfoil Generator (AG) outperforms all other parameterization methods with the same number of design variables. Additionally, $\text{AG}_{8}$ is comparable to 
$\text{CST}_{10}$, $\text{B\'ezier}_{10}$, and $\text{PAREC}_{10}$ even though it has a smaller number of design variables.

\begin{figure*}[htb!]
	\centering
		\includegraphics[width=1.\textwidth]{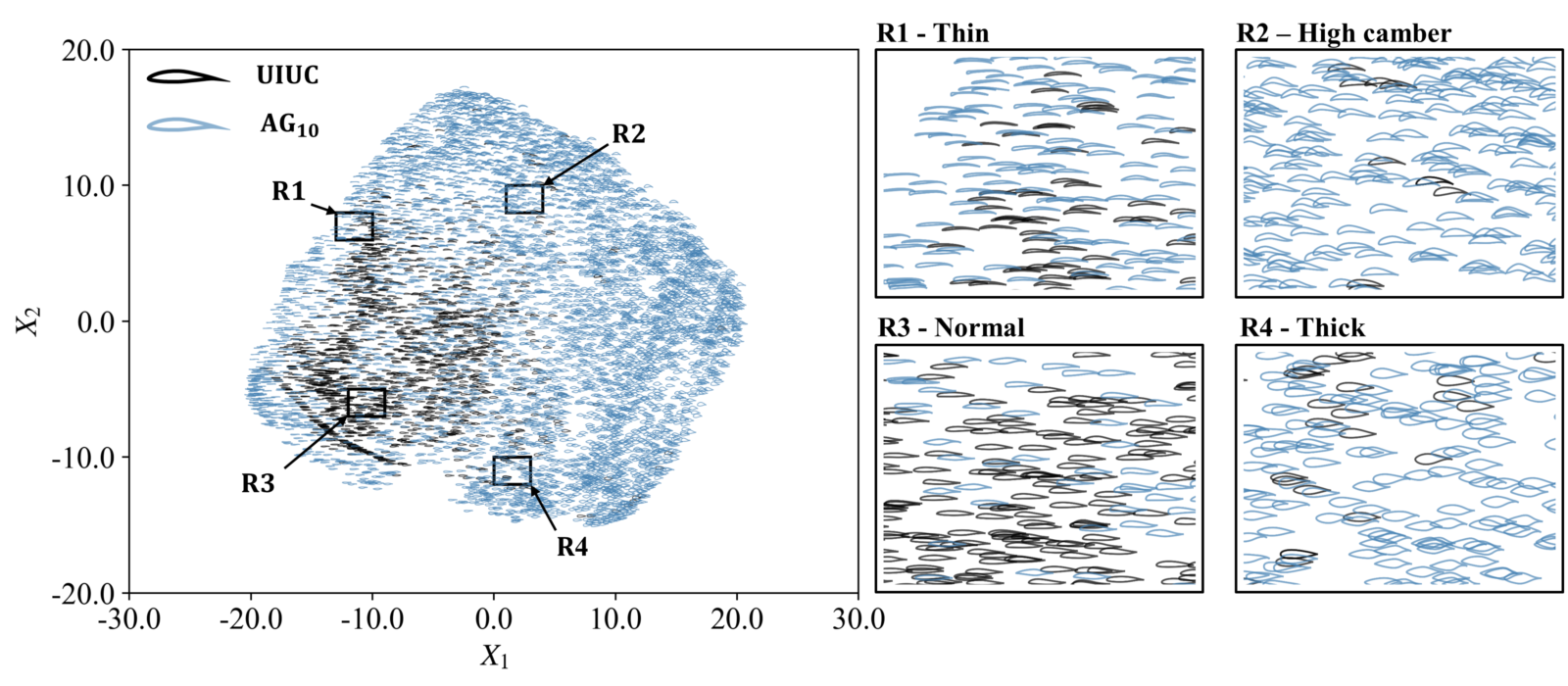}
	\caption{$t$-SNE visualization of the design space of Airfoil Generator.}
	\label{fig:tsne}
\end{figure*}

To visualize the design space of the Airfoil Generator, a $t$-SNE plot was used and the result is shown in Fig. \ref{fig:tsne}. The airfoils randomly generated from $\text{AG}_{10}$ and the airfoils from UIUC database are plotted in two-dimensional manifold learned by the $t$-SNE algorithm, which is represented as the $X_1-X_2$ plane in Fig. \ref{fig:tsne}. Four regions are highlighted based on their geometric characteristics: thin (R1), high camber (R2), normal (R3), and thick (R4). It can be seen that the airfoils generated by $\text{AG}_{10}$ largely cover the UIUC database. We believe that the impressive generalizability of the Airfoil Generator stems from two key aspects: first, since the structure of the Airfoil Generator explicitly decouples the generation of the thickness and camber distributions, it can effectively permute the thickness and camber of existing airfoils, greatly expanding the design space. Secondly, within the Airfoil Generator, the physical latent dimensions correlate directly with the physical attributes of airfoils, including maximum thickness/camber, trailing edge angle, and leading edge radius. This setup facilitates the independent manipulation of the physical attributes, even though they are inherently related (e.g. maximum thickness and leading edge radius are positively correlated, as confirmed in Fig. \ref{fig:Visualization_OOD}). This characteristic further expands the design space of the Airfoil Generator.

\subsection{Feasibility and Intuitiveness}\label{sec:feasibility_intuitiveness}
\begin{figure}[htb!]
	\centering
	
	\begin{subfigure}[b]{0.475\linewidth}
        \centering
        \includegraphics[width=\linewidth]{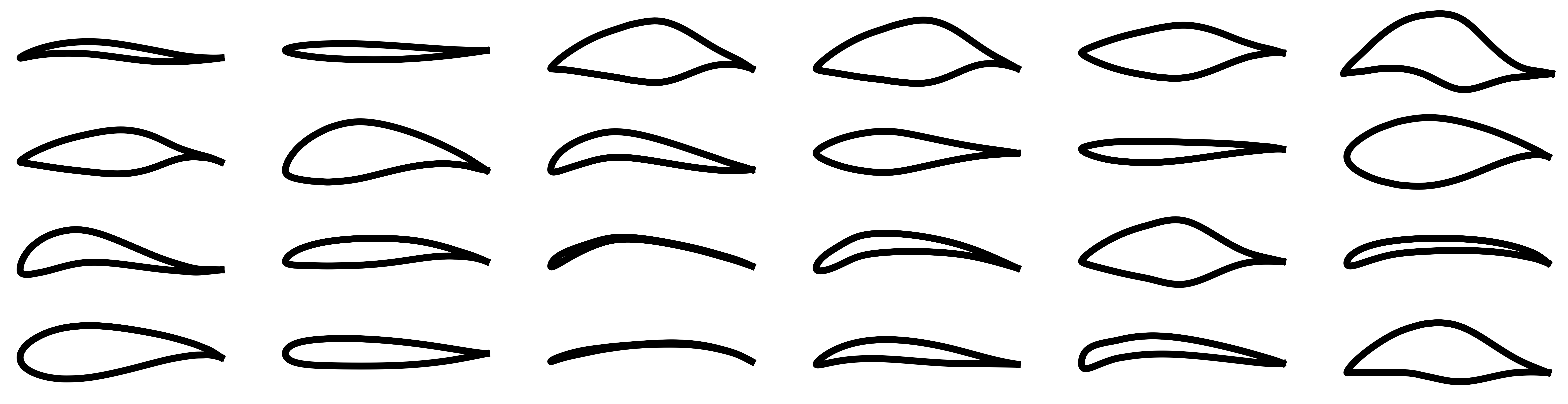}
        \caption{Airfoil Generator (AG)}
        \label{fig:random_sample_ag10}
    \end{subfigure}
    \hfill 
    \begin{subfigure}[b]{0.475\linewidth}
        \centering
        \includegraphics[width=\linewidth]{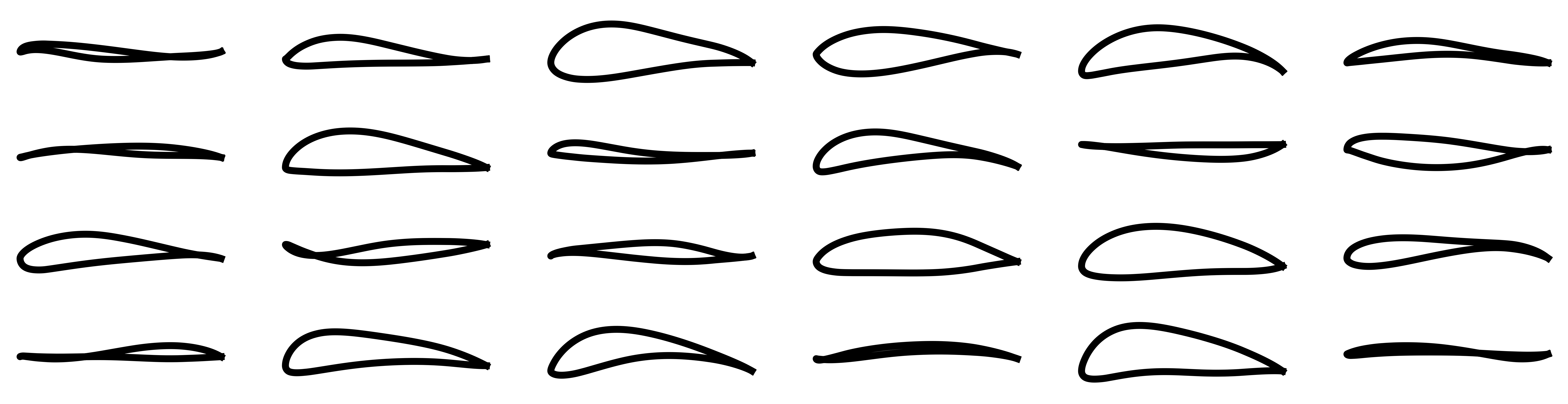}
        \caption{Singular value decomposition (SVD)}
        \label{fig:random_sample_svd10}
    \end{subfigure}

    \vspace{1em} 
    
    \begin{subfigure}[b]{0.475\linewidth}
        \centering
        \includegraphics[width=\linewidth]{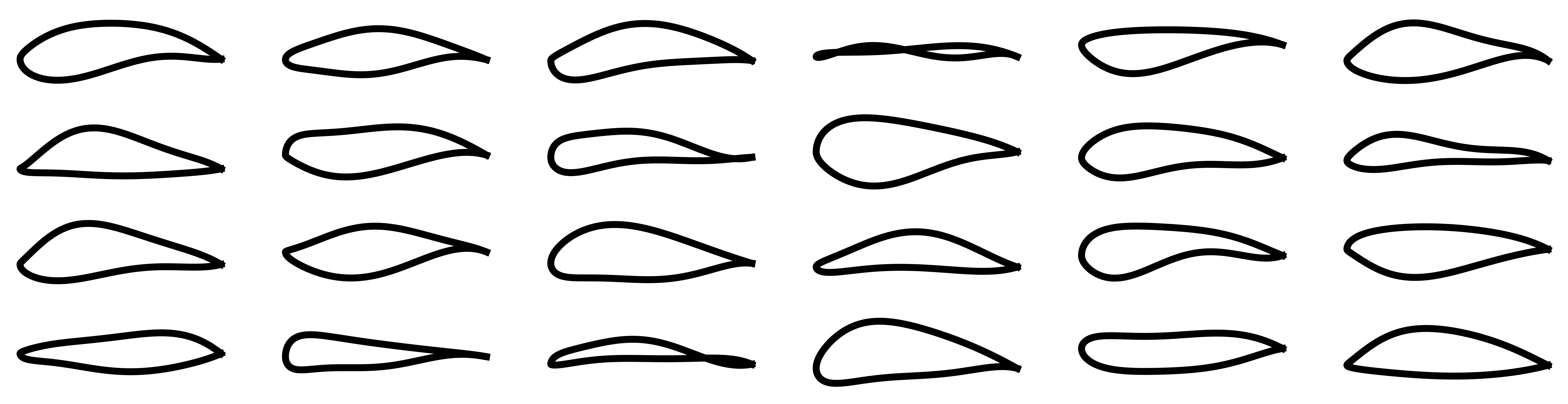}
        \caption{Class/shape-function transformation (CST)}
        \label{fig:random_sample_cst10}
    \end{subfigure}
    \hfill 
    \begin{subfigure}[b]{0.475\linewidth}
        \centering
        \includegraphics[width=\linewidth]{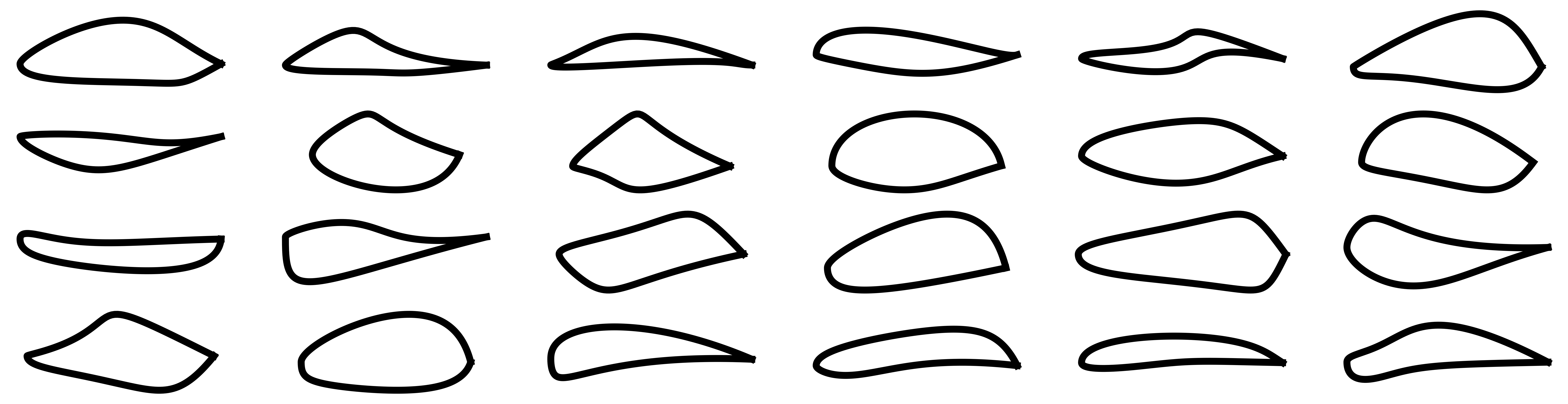}
        \caption{B\'ezier}
        \label{fig:random_sample_bezier10}
    \end{subfigure}
    
    \caption{Random airfoil generation for each parameterization method}
    \label{fig:random_sampling}
\end{figure}

To investigate the feasibility of each method, airfoils that randomly generated from each parameterization method are plotted in Fig. \ref{fig:random_sampling}. The sampling range is the same with that of the inverse fitting was conducted. It can be confirmed that the Airfoil Generator well-produce feasible airfoils, the geometries of which are smooth and non-intersecting, while other methods generate self-intersecting airfoils. Table \ref{tab:FR_intersect} quantitatively compares the feasibility ratio, which is defined by the number of non-intersecting airfoils divided by the total number of sampled airfoils. While other methods produce 20-40$\%$ of self-intersecting airfoils, AG and IGP did not produce self-intersecting airfoils at all.

\begin{table}[htb!]
\centering
\setlength\tabcolsep{20pt} 
\caption{Feasibility ratio of randomly generated airfoils; constraint on self-intersecting geometry.}
\label{tab:FR_intersect}
\begin{tabular}{lcc}
\hline \hline
Methods & $N_\text{DV}$ & Feasibility ratio [\%] \\
\hline
\multirow{2}{*}{AG} & 10 & 100.0 \\
& 8 & 100.0 \\
\multirow{2}{*}{SVD} & 10 & 57.7 \\
& 8 & 60.4 \\
\multirow{2}{*}{CST} & 10 & 69.8 \\
& 8 & 72.7 \\
B\'ezier & 10 & 83.1 \\
PARSEC & 10 & 51.4 \\
IGP & 8 & 100.0 \\
\hline \hline
\end{tabular}
\end{table}

For intuitiveness, we defined the quantitative criteria to measure intuitiveness: the maximum Pearson correlation coefficient, $C_\text{max}$. First, Pearson correlation coefficients between design variables and physical features were calculated. Next, the Pearson correlation coefficient of the design variables with the maximum value is selected as the representative value. This criterion can be interpreted as how the single design variable is well-aligned with the physical feature. The results are shown in Table. \ref{tab:max_pearson}. It is confirmed that $C_\text{max}$ of AG reaches 1 in all physical features. This is because the latent variables of AG are trained to be directly aligned with the physical variables, as discussed in Fig. \ref{fig:CV_phys}. On the contrary, the $C_\text{max}$ of other parametrization methods are far below that of AG.

\begin{table}[htb!]
\centering
\setlength\tabcolsep{20pt} 
\caption{Maximum Pearson correlation coefficients for each physical features.}
\label{tab:max_pearson}
\begin{tabular}{lcccc}
\hline \hline
\multirow{2}{*}{Methods} & \multicolumn{4}{c}{$C_\text{max}$}\\
& \( m_{\text{max}} \) & \( \gamma_{\text{TE}} \) & \( t_{\text{max}} \) & \( \log(r_{\text{LE}}) \) \\
\hline
AG & \textbf{1.00} & \textbf{0.99} & \textbf{1.00} & \textbf{1.00} \\
SVD & 0.76 & 0.54 & 0.82 & 0.65 \\
CST & 0.47 & 0.72 & 0.57 & 0.76 \\
B\'ezier & 0.53 & 0.46 & 0.52 & 0.54 \\
IGP & 0.75 & 0.75 & \textbf{0.99} & 0.83 \\
\hline \hline
\end{tabular}
\end{table}

\begin{figure*}[htb!]
	\centering
\includegraphics[width=1.\textwidth]{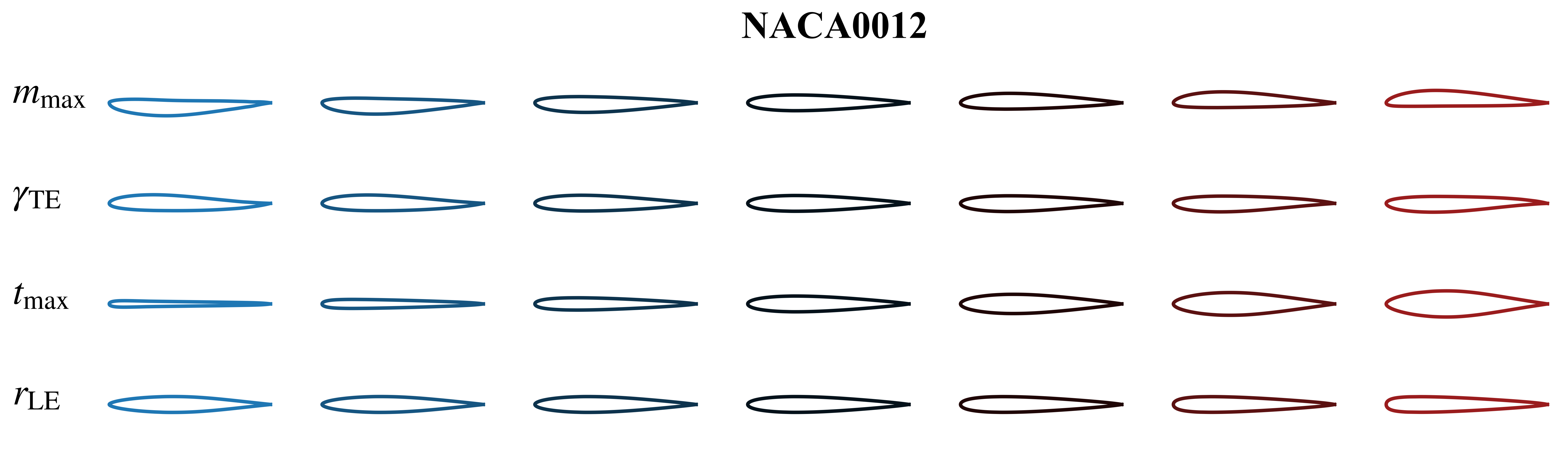}
	\caption{Latent traversal plot of physical latent variables in Airfoil Generator.}
	\label{fig:LV_sweep}
\end{figure*}

\begin{figure}[htb!]
	\centering
	\begin{subfigure}[b]{0.8\linewidth}
        \centering
        \includegraphics[width=\linewidth]{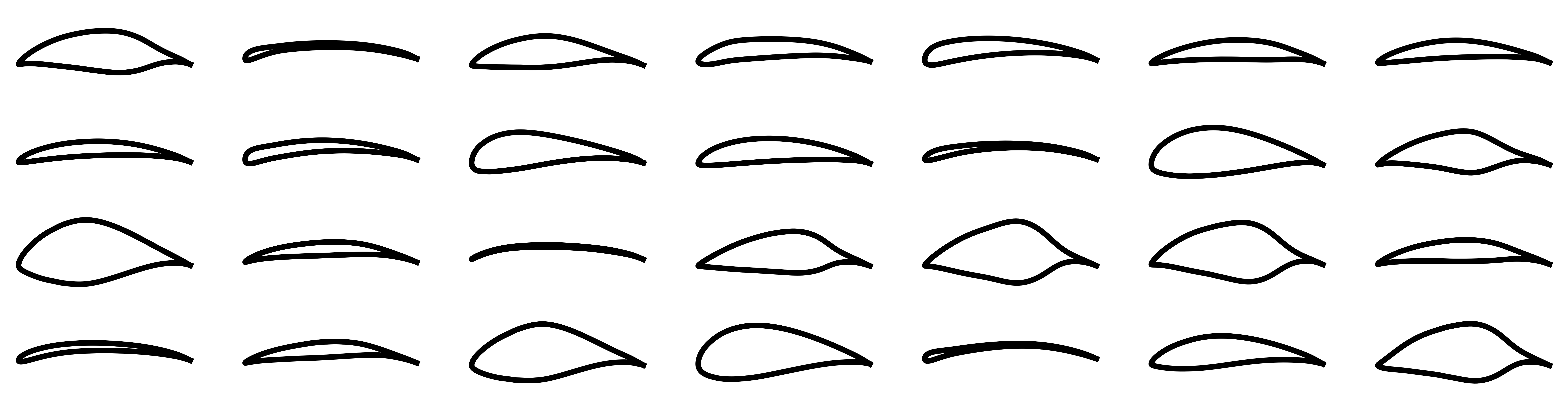}
        \caption{With the fixed maximum camber and trailing edge angle}
        \label{fig:constraint_random_gen_2}
    \end{subfigure}
    \vspace{1em} 

    \begin{subfigure}[b]{0.8\linewidth}
        \centering
        \includegraphics[width=\linewidth]{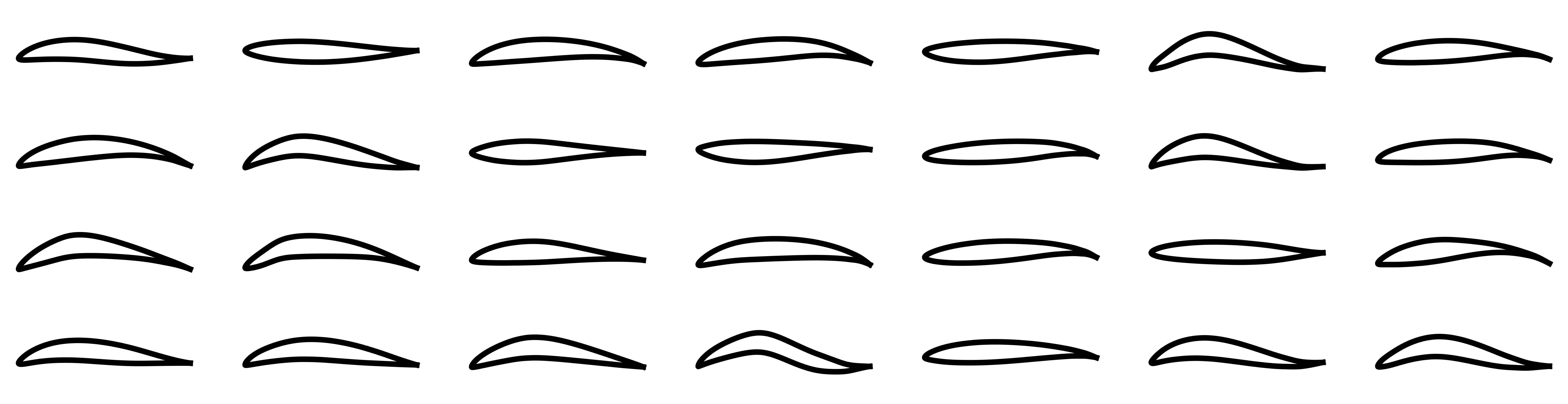}
        \caption{With the fixed maximum thickness and leading edge radius}
        \label{fig:constraint_random_gen_1}
    \end{subfigure}
    \vspace{1em} 
    \caption{Random airfoil generation using Airfoil Generator with fixed physical latent variables}
    \label{fig:constraint_random_gen}
\end{figure}

To qualitatively confirm the intuitiveness of AG, the effect of each physics latent variable is visualized by a latent traversal plot, as shown in Fig. \ref{fig:LV_sweep}. The baseline airfoil was selected as NACA0012, and each physics latent variable was swept while other latent variables remained fixed. It is confirmed that each physics latent variable controls the physical property of interest. This type of ``direct mapping'' between design variables and physical features benefits the user by allowing direct modification of airfoil attributes, and also simplifies the imposition of physical constraints. Figure \ref{fig:constraint_random_gen} shows an example of constraint imposition using physical latent variables. The random airfoil samples are generated with the fixed maximum camber and trailing edge angle (Fig. \ref{fig:constraint_random_gen_2}) and with the fixed maximum thickness and leading edge radius (Fig. \ref{fig:constraint_random_gen_1}, while other attributes are free to change. It can be observed that a variety of design candidates can be sampled while the specified physical features remain fixed. This capability significantly improves the efficiency of sampling within physical constraints, which is highly practical for the constrained airfoil shape optimization. This aspect is discussed in more detail in the following section.

\subsection{Airfoil Shape Optimization}\label{sec:ASO}
In order to investigate the combined effect of the desirable properties discussed so far, airfoil shape optimizations have been performed. Two benchmark optimization problems are selected, one for unconstrained and one for constrained. The purpose of the unconstrained optimization is to investigate the flexibility of each parameterization method by pushing it to an extreme situation. Although the optimization results may converge to an unrealistic airfoil shape, it can effectively demonstrate the flexibility of the parameterization methods. On the other hand, constrained optimization, which is a more common case for practical application, is selected to investigate the effectiveness of physically intuitive parameterization. 

The specifications for each optimization problem are presented in Table. \ref{tab:optimization_def}. The objective function of both optimization problems is the lift-to-drag ratio. In the unconstrained optimization, however, the only constraint ensures a non-intersecting geometry, which is given by $t_\text{min} \geq 0$. On the other hand, constrained optimization imposes constraints on the maximum thickness, the maximum camber, and the leading edge radius. 

\begin{table}[htb!]
\centering
\caption{Specifications for unconstrained and constrained optimization problem.}
\label{tab:optimization_def}
\setlength\tabcolsep{20pt} 
\begin{tabular}{lcc}
\hline \hline
& \textbf{Unconstrained} & \textbf{Constrained} \\
\hline
\textbf{Maximize} & \( C_l/C_d \) & \( C_l/C_d \) \\
\hline
\textbf{Subject to} & \( t_{\text{min}} > 0 \) & \( t_{\text{min}} > 0 \) \\
& & \( m_{\text{max}} < 0.03 \) \\
& & \( 0.1 < t_{\text{max}} < 0.12 \) \\
& & \( r_{\text{LE}} < 0.005 \) \\
\hline \hline
\end{tabular}
\end{table}

Here are the details of the imposition of physical constraints in the constrained optimization problem: First, for AG and IPG, which have design variables directly related to physical features, the search domain of the corresponding design variables is properly adjusted (it is worth noting that although PARSEC's design variables are also correlated with physical features such as upper/lower crest locations, but they do not have a direct relationship with the constraints of interest). In contrast, for other methods, where the design variables are not directly related to physical features, the airfoil was generated first, its physical properties were calculated numerically, and then the constraints were applied.

Within the physical constraints, direct optimization was performed using XFOIL \cite{drela1989xfoil}. The flow conditions in XFOIL were set to $Re = 1.8\times10^6$, $Ma = 0.01$, and $AoA = 0^{\circ}$, consistent with ref. \cite{chen2020airfoil}. A genetic algorithm was employed as the global optimizer, with both population and generation set to 100, resulting in a total of 10,000 evaluations for each optimization case. Since the optimization results can vary significantly depending on the initial population distribution, the process was repeated ten times. The mean and variance of the objective function are then analyzed.

\begin{figure*}[htb!]
	\centering
\includegraphics[width=1.\textwidth]{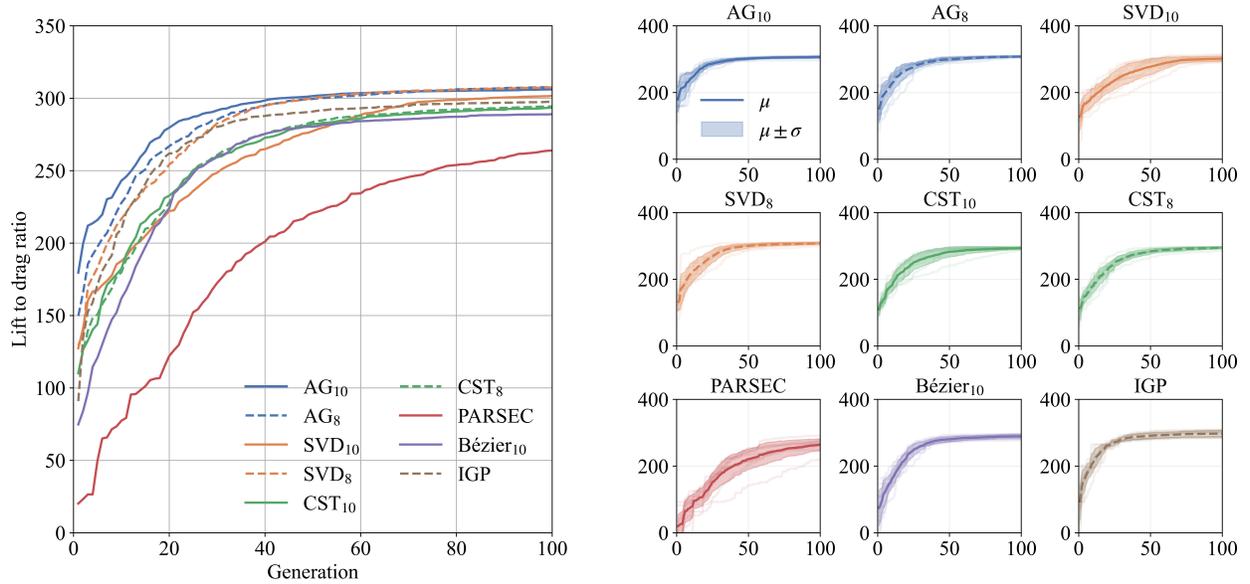}
	\caption{Unconstrained optimization results; convergence history.}
	\label{fig:unconstrained_opt_res}
\end{figure*}

\begin{figure*}[htb!]
	\centering
\includegraphics[width=0.65\textwidth]{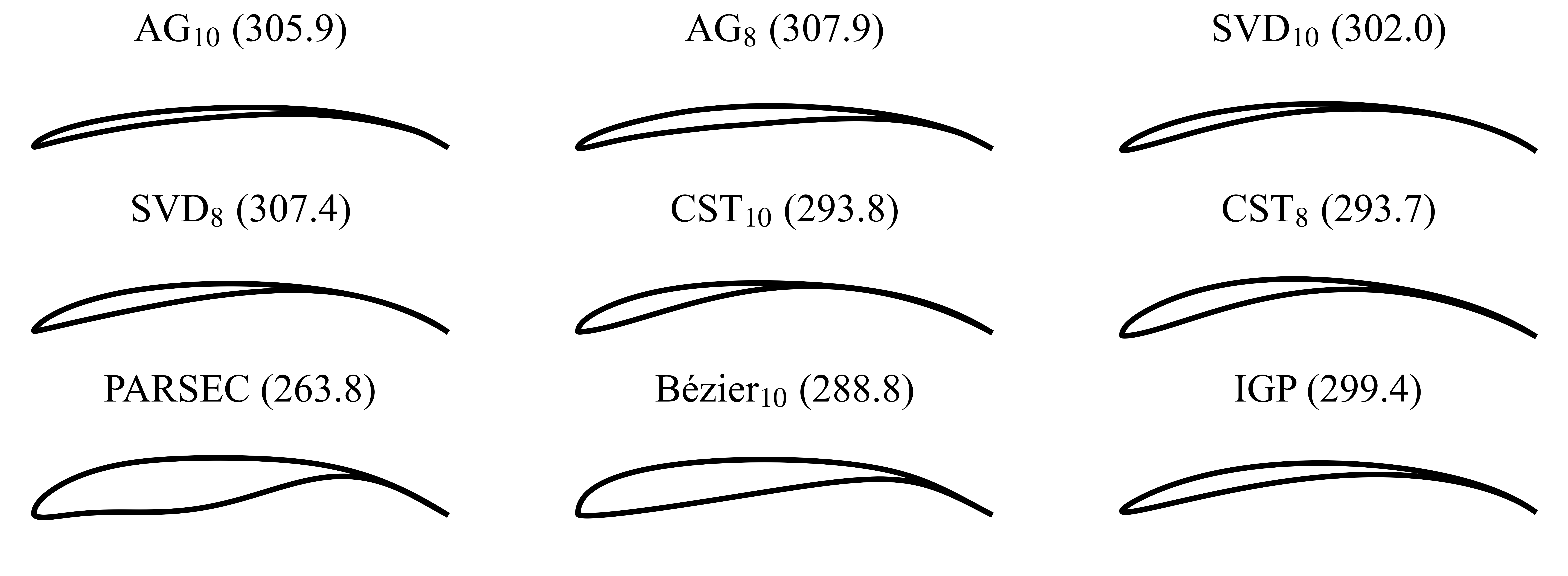}
	\caption{Unconstrained optimization results; optimal airfoil.}
	\label{fig:unconstraint_opt_res_airfoil}
\end{figure*}

\begin{figure*}[htb!]
	\centering
\includegraphics[width=1.\textwidth]{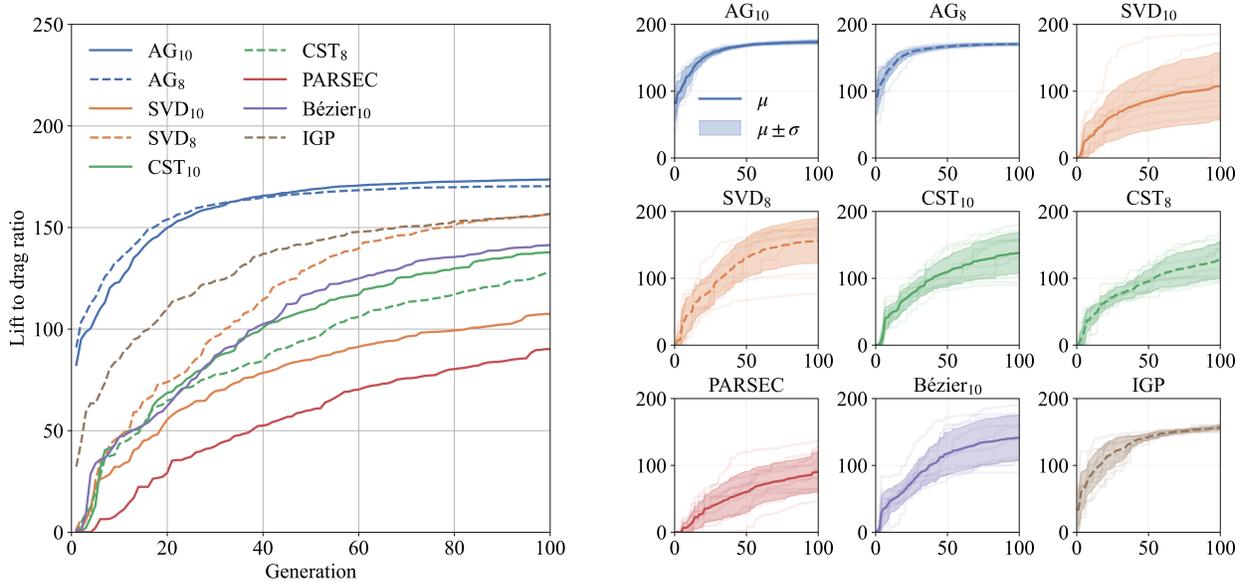}
	\caption{Constrained optimization results; convergence history.}
	\label{fig:constrained_opt_res}
\end{figure*}

\begin{figure*}[htb!]
	\centering
\includegraphics[width=0.7\textwidth]{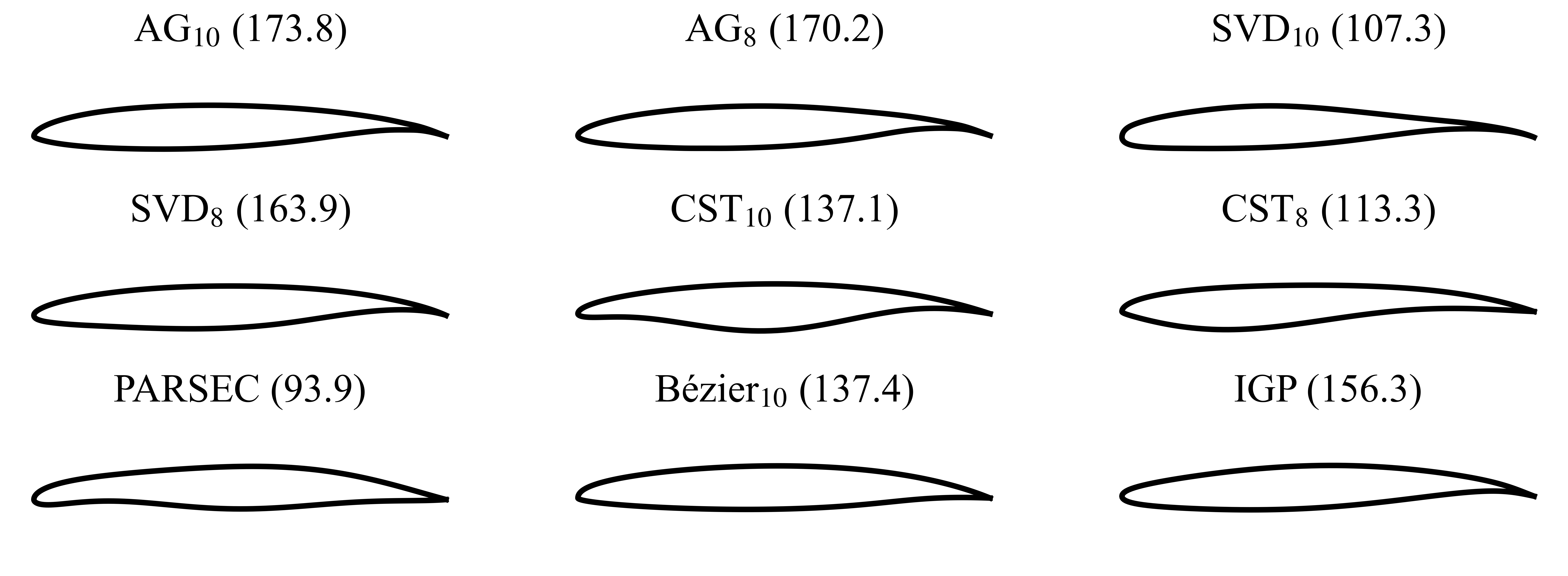}
	\caption{Constrained optimization results; optimal airfoil.}
	\label{fig:constraint_opt_res_airfoil}
\end{figure*}

The optimization results are visualized in Figs. \ref{fig:unconstrained_opt_res}-\ref{fig:constraint_opt_res_airfoil}. First, Figs. \ref{fig:unconstrained_opt_res} and \ref{fig:constrained_opt_res} represent the optimization history. On the right side of the figure, where there are nine sub-figures, the series of optimization results for each parameterization technique are presented. Within each sub-figure, the lines with reduced opacity represent the individual results of each optimization run, while the fully opaque line represents the mean. The shaded area shows the one-sigma confidence interval. On the left side of the figure, the mean histories corresponding to all parameterization methods are plotted together. The optimized airfoil for each parameterization method is shown in figures \ref{fig:unconstraint_opt_res_airfoil} and \ref{fig:constraint_opt_res_airfoil}. In Figs. \ref{fig:unconstraint_opt_res_airfoil} and \ref{fig:constraint_opt_res_airfoil}, the optimized airfoils for each parameterization method are displayed, with the representative case whose converged results are closest to the mean of ten repeated trials.

On the right side of Fig. \ref{fig:unconstrained_opt_res}, the ten optimization trials have similar histories within a narrow confidence interval. It is interesting to note that the random initialization has minimal impact on the final results. On the left, the average history shows that all methods exceeded an optimal lift-to-drag value of 285 (cf. AG and SVD even exceeded 300), with the sole exception of PARSEC, which reached 263. Within Fig. \ref{fig:unconstraint_opt_res_airfoil}, a consistent trend can be seen across most parameterization methods: optimized airfoils typically have an extremely thin thickness, significantly high values for both camber and trailing edge angle. These characteristics explain the suboptimal performance of the PARSEC. In PARSEC, a linear system needs to be solved to correlate the polynomial coefficient with the physical features of the airfoil. However, it is known that this linear system becomes ill-conditioned when the maximum camber location is close to zero \cite{wu2003comparisons}, which is the case in this problem. Such a situation highlights a fundamental limitation of polynomial-based parameterizaiton methods: efforts to directly link design variables to physical features can result in a design space that is ill-conditioned, thereby reducing feasibility.

For the unconstrained optimization problem, aspects such as convergence, reliability, and optimal value are comparable among the parameterization methods, with a small exception of PARSEC. However, when it comes to the constrained optimization problem, completely different results were observed. As can be seen on the right side of the figure \ref{fig:constrained_opt_res}, the confidence intervals are much wider compared to those in the unconstrained optimization. Except for the AG and IGP methods, where the physical constraints were handled directly by physically intuitive design variables, the optimization results varied significantly from test run to test run, making the optimization results unreliable. Additionally, the overall convergence speed was significantly slowed down, and the resulting optimal value is much smaller than that of AG. 

\begin{figure}[H]
	\centering
	
	\begin{subfigure}[b]{.8\linewidth}
        \centering
        \includegraphics[width=\linewidth]{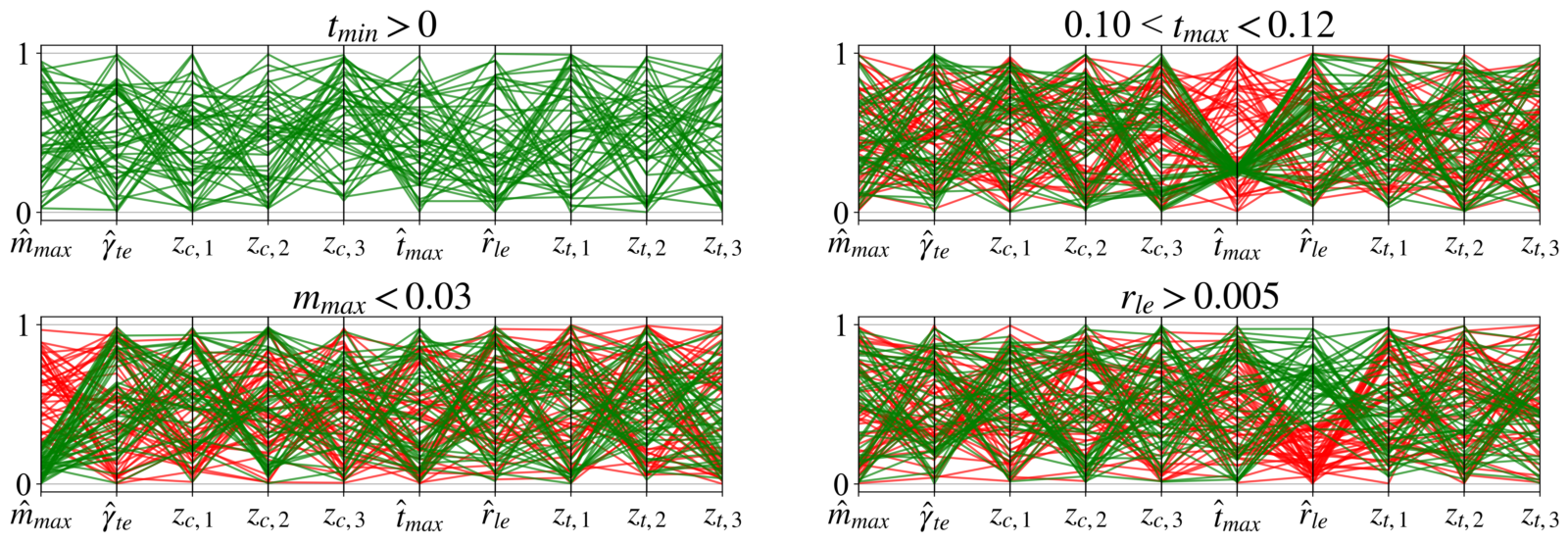}
        \caption{Airfoil Generator (AG)}
        \label{fig:AG10_DS_vis}
    \end{subfigure}
    
    \vspace{1em} 

    \begin{subfigure}[b]{.8\linewidth}
        \centering
        \includegraphics[width=\linewidth]{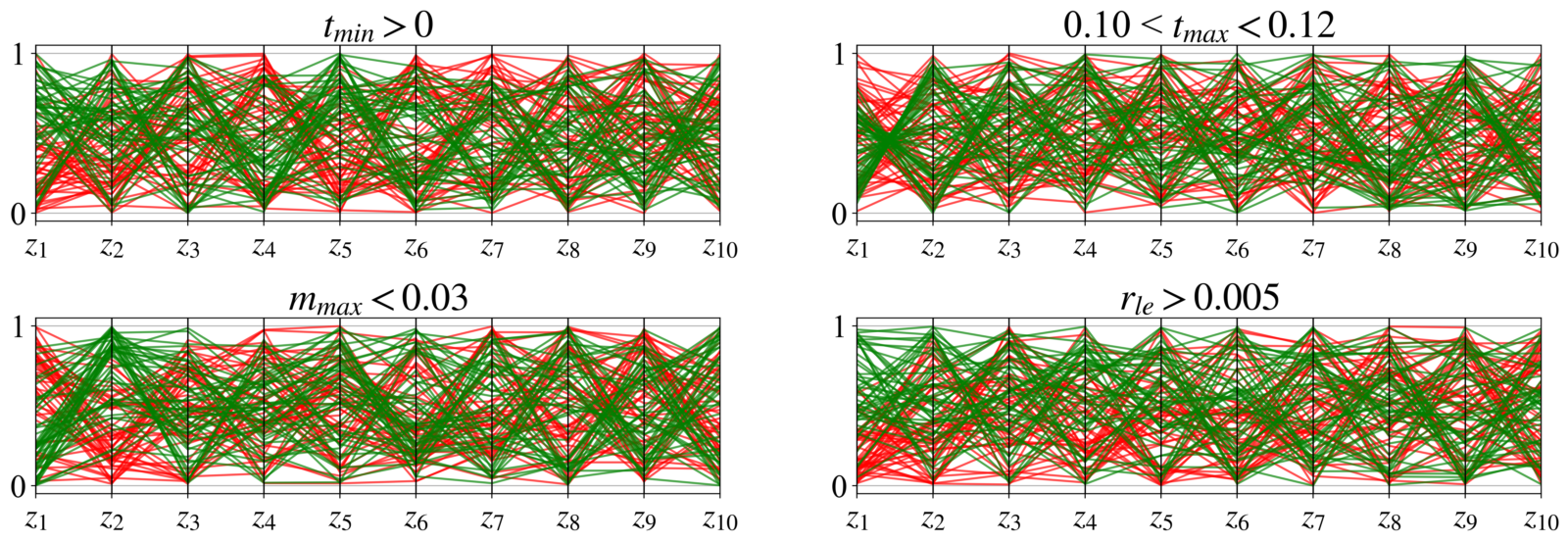}
        \caption{Singular value decomposition (SVD)}
        \label{fig:SVD10_DS_vis}
    \end{subfigure}
    
    \vspace{1em} 

    \begin{subfigure}[b]{.8\linewidth}
        \centering
        \includegraphics[width=\linewidth]{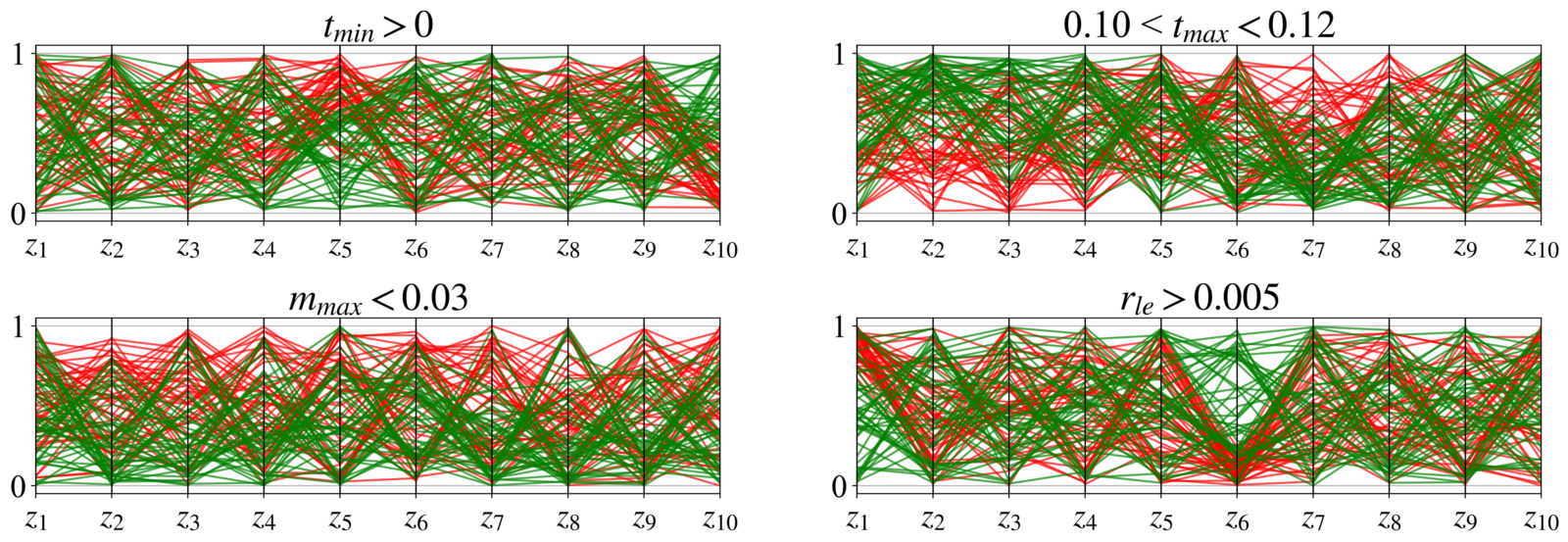}
        \caption{Class/shape-function transformation (CST)}
        \label{fig:CST10_DS_vis}
    \end{subfigure}

    \vspace{1em} 

    \begin{subfigure}[b]{.8\linewidth}
        \centering
        \includegraphics[width=\linewidth]{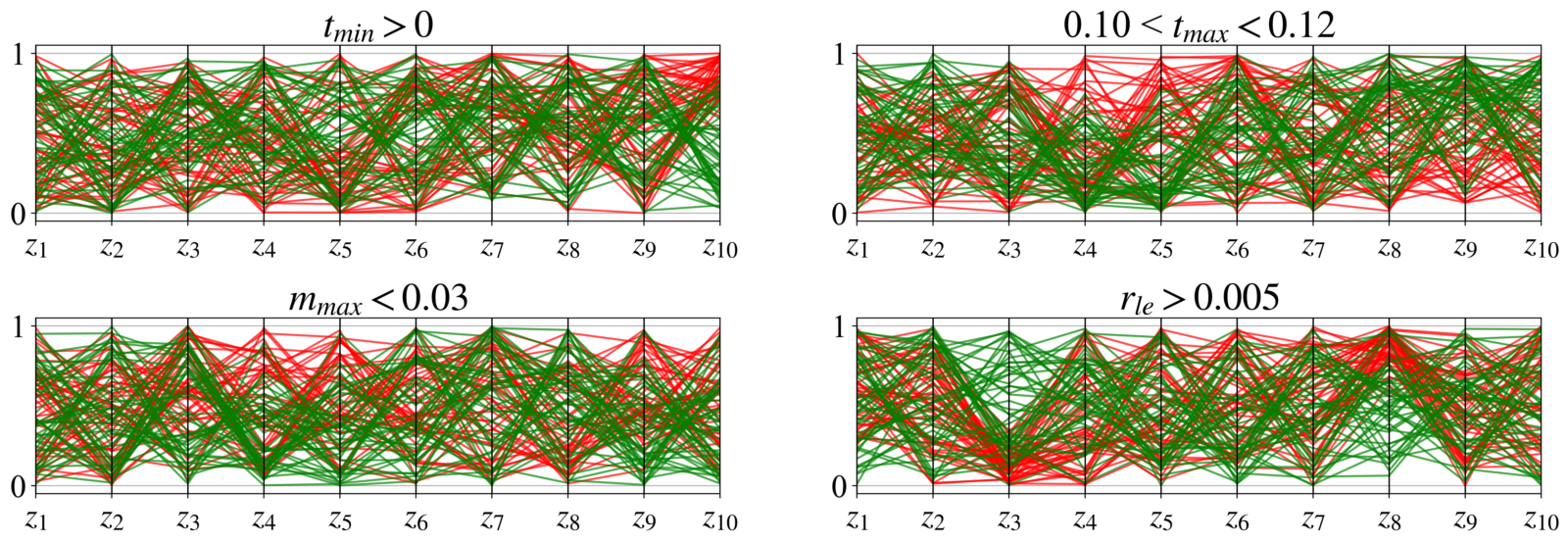}
        \caption{B\'ezier}
        \label{fig:Bezier10_DS_vis}
    \end{subfigure}
    
    \caption{Visualization of constrained design space using parallel coordinates}
    \label{fig:constrained_ds_vis}
\end{figure}

The authors hypothesize that the imposition of a physical constraint through post-processing may have fragmented the design space into multiple localized zones. This could increase the dependence on initialization and slow down the convergence rate. To verify this conjecture, we visualized the constrained design spaces for $\text{AG}_{10}$, $\text{SVD}_{10}$, $\text{CST}_{10}$, $\text{B\'ezier}_{10}$, using parallel coordinates, as shown in Fig. \ref{fig:constrained_ds_vis}. Each figure is composed of four sub-figures, each illustrating the effect of a specific physical physical constraint: $t_\text{min}>0$, $0.1 < t_\text{max} < 0.12$, $m_\text{max}<0.03$, and $r_\text{LE}>0.005$. Within these plots, each vertical line represents a specific design variable, which has been normalized to range between 0 and 1. The interconnected lines traversing these vertical axes represent individual airfoil designs. Green colored lines represent designs that meet the specified constraints, while red colored lines represent designs that do not meet the specified constraints.

Figure \ref{fig:AG10_DS_vis} shows the advantages of the Airfoil Generator over other parameterization methods. First, as previously highlighted in Table. \ref{tab:FR_intersect}, the Airfoil Generator consistently satisfies the non-intersecting constraints, regardless of the combination of design variables. Second, due to the direct alignment of physical features with the design variables - denoted as $\hat{m}_\text{max}$, $\hat{\gamma}_\text{TE}$, $\hat{t}_\text{max}$, and $\hat{r}_\text{LE}$ - the imposition of physical constraints is both independent and exclusive. This means that an individual constraint predominantly affects its corresponding design variable (independent), and crucially, the regions where design constraints are violated are clearly separated from those that are compliant (exclusive). As a result of these properties, the imposition of physical constraints can be achieved efficiently by adjusting the range of the design variable.

On the other hand, these desirable properties are not evident in other parameterization methods, as can be seen in Figs. \ref{fig:SVD10_DS_vis}-\ref{fig:Bezier10_DS_vis}. Although there are a discernible correlations between the constraints and the design variables, they are neither independent nor exclusive. For example, for the samples that satisfy the maximum thickness constraint, $z_1$ and $z_2$ have an inversely proportional relationship in SVD, $z_{2-3}$ and $z_{7}$ in CST, and $z_{4-5}$ and $z_{9}$ in B\'ezier. However, due to the inherent ambiguity of these correlations, it is difficult to precisely define the feasible range. Moreover, even if the range can be defined, it does not guarantee the elimination of all nonconforming samples.

In summary, for unconstrained optimization, the convergence rate and the reliability of the optimization results were acceptable in most parameterization methods. However, for constrained optimization, physical constraints that are not directly related to the design variables complicate the exploration of the design space, degrading both efficiency and reliability. The use of intuitive parameterization methods, where design variables are directly related to physical features, can alleviate this problem by clearly defining the feasible design space.

\section{Conclusion}\label{sec:conclusion}

In this study, a compact and intuitive airfoil parameterization method, namely Airfoil Generator, has been developed using physics-aware variational autoencoder. Due to the flexibility inherent in the neural network, Airfoil Generator can achieve compact representation of large airfoil databases while maintaining a sufficient level of flexibility. Furthermore, due to the novel structure of Airfoil Generator proposed in this study, which not only decomposes the generation of thickness and camber distribution, but also directly feeds back the latent dimension to align with the physical features, a strong intuitiveness is achieved, which enables the direct imposition of physical constraints.

The superiority of the Airfoil Generator is verified by an extensive comparative study with different parameterization methods, such as parameterized section (PARSEC), class/shape function transformation (CST), singular value decomposition (SVD), improved geometric parameterization (IGP), and B\'ezier curve. The results of the comparative study can be highlighted as follows:

\textit{\textbf{Parsimony and Flexibility}}: The flexibility of each parameterization is compared through an inverse fitting of 300 randomly sampled airfoils from the UIUC airfoil database. When compared with the same number of design variables, the Airfoil Generator outperformed all other parameterization methods in terms of mean squared error between target and fitted airfoils. It is noteworthy that the flexibility of Airfoil Generator with the 8 design variables is comparable to the CST method with the 10 design variables.

\textit{\textbf{Feasibility and Intuitiveness}}: The feasibility of each parameterization is compared by the feasibility ratio, which measures the constraint satisfaction rate of randomly generated airfoils. The Airfoil Generator achieved a feasibility ratio of 100 \% in terms of non-intersecting criteria, while the other parameterization methods were below 80 \%, with the exception of IGP. Additionally, the intuitiveness is compared using both qualitative and quantitative measures, confirming superiority of Airfoil Generator. 

Airfoil shape optimizations were also carried out to evaluate the overall benefits of the above desirable properties. Both unconstrained and constrained optimizations were performed, and it was confirmed that Airfoil Generator outperformed other parameterization methods in terms of convergence rate and search capability. In particular, in the constrained optimization problem, the Airfoil Generator shows an exceptionally high reliability in repeated optimization runs. It was shown that a physical constraint that is not aligned with the design variable can significantly degrade the convergence of the optimization results, highlighting the importance of aligning the design variables with the physical features.

Our research group is currently integrating the Airfoil Generator model into the design optimization framework for rotor blades, and is investigating how the optimization efficiency can be improved compared to the conventional optimization framework. However, the question of which physical features of the airfoil are optimal for training the Airfoil Generator, and how the selection of other generative model families, such as the diffusion model, can affect the model efficiency, remains unresolved and is highly desirable to be investigated in future studies.



\bibliography{sample}

\begin{thebibliography}{46}
\newcommand{\enquote}[1]{``#1''}
\providecommand{\natexlab}[1]{#1}
\providecommand{\url}[1]{\texttt{#1}}
\providecommand{\urlprefix}{URL }
\expandafter\ifx\csname urlstyle\endcsname\relax
  \providecommand{\doi}[1]{\discretionary{}{}{}https://doi.org/#1}\else
  \providecommand{\doi}[1]{\discretionary{}{}{}\urlstyle{rm}\url{https://doi.org/#1}}\fi

\bibitem[{Cook and Jarrett(2017)}]{cook2017robust}
Cook, L.~W., and Jarrett, J.~P., \enquote{Robust airfoil optimization and the importance of appropriately representing uncertainty,} \emph{AIAA Journal}, Vol.~55, No.~11, 2017, pp. 3925--3939.

\bibitem[{Massaro and Benini(2012)}]{massaro2012multi}
Massaro, A., and Benini, E., \enquote{Multi-objective optimization of helicopter airfoils using surrogate-assisted memetic algorithms,} \emph{Journal of Aircraft}, Vol.~49, No.~2, 2012, pp. 375--383.

\bibitem[{Samareh(2001)}]{samareh2001survey}
Samareh, J.~A., \enquote{Survey of shape parameterization techniques for high-fidelity multidisciplinary shape optimization,} \emph{AIAA Journal}, Vol.~39, No.~5, 2001, pp. 877--884.

\bibitem[{Wu et~al.(2003)Wu, Yang, Liu, and Tsai}]{wu2003comparisons}
Wu, H.-Y., Yang, S., Liu, F., and Tsai, H.-M., \enquote{Comparisons of three geometric representations of airfoils for aerodynamic optimization,} \emph{16th AIAA Computational Fluid Dynamics Conference}, Orlando, Florida, 2003.

\bibitem[{Song and Keane(2004)}]{song2004study}
Song, W., and Keane, A., \enquote{A study of shape parameterisation methods for airfoil optimisation,} \emph{10th AIAA/ISSMO Multidisciplinary Analysis and Optimization Conference}, Albany, New York, 2004.

\bibitem[{Sripawadkul et~al.(2010)Sripawadkul, Padulo, and Guenov}]{sripawadkul2010comparison}
Sripawadkul, V., Padulo, M., and Guenov, M., \enquote{A comparison of airfoil shape parameterization techniques for early design optimization,} \emph{13th AIAA/ISSMO Multidisciplinary Analysis Optimization Conference}, Fort Worth, Texas, 2010.

\bibitem[{Zhu and Qin(2014)}]{zhu2014intuitive}
Zhu, F., and Qin, N., \enquote{Intuitive class/shape function parameterization for airfoils,} \emph{AIAA Journal}, Vol.~52, No.~1, 2014, pp. 17--25.

\bibitem[{Masters et~al.(2017)Masters, Taylor, Rendall, Allen, and Poole}]{masters2017geometric}
Masters, D.~A., Taylor, N.~J., Rendall, T., Allen, C.~B., and Poole, D.~J., \enquote{Geometric comparison of aerofoil shape parameterization methods,} \emph{AIAA Journal}, Vol.~55, No.~5, 2017, pp. 1575--1589.

\bibitem[{Sobieczky(1999)}]{sobieczky1999parametric}
Sobieczky, H., \enquote{Parametric airfoils and wings,} \emph{Recent Development of Aerodynamic Design Methodologies: Inverse Design and Optimization}, Springer, 1999, pp. 71--87.

\bibitem[{Kulfan(2008)}]{kulfan2008universal}
Kulfan, B.~M., \enquote{Universal parametric geometry representation method,} \emph{Journal of Aircraft}, Vol.~45, No.~1, 2008, pp. 142--158.

\bibitem[{Piegl and Tiller(1996)}]{piegl1996nurbs}
Piegl, L., and Tiller, W., \emph{The NURBS book}, Springer Science \& Business Media, 1996.

\bibitem[{Toal et~al.(2010)Toal, Bressloff, Keane, and Holden}]{toal2010geometric}
Toal, D.~J., Bressloff, N.~W., Keane, A.~J., and Holden, C.~M., \enquote{Geometric filtration using proper orthogonal decomposition for aerodynamic design optimization,} \emph{AIAA Journal}, Vol.~48, No.~5, 2010, pp. 916--928.

\bibitem[{Derksen and Rogalsky(2010)}]{derksen2010bezier}
Derksen, R., and Rogalsky, T., \enquote{Bezier-PARSEC: An optimized aerofoil parameterization for design,} \emph{Advances in Engineering Software}, Vol.~41, No. 7-8, 2010, pp. 923--930.

\bibitem[{Lu et~al.(2018)Lu, Huang, Song, and Li}]{lu2018improved}
Lu, X., Huang, J., Song, L., and Li, J., \enquote{An improved geometric parameter airfoil parameterization method,} \emph{Aerospace Science and Technology}, Vol.~78, 2018, pp. 241--247.

\bibitem[{Selig(1996)}]{selig1996uiuc}
Selig, M., \enquote{UIUC Airfoil Data Site,} University of Illinois Urbana-Champaign, 1996.
\newblock \urlprefix\url{https://m-selig.ae.illinois.edu/ads.html}.

\bibitem[{Bengio et~al.(2013)Bengio, Courville, and Vincent}]{bengio2013representation}
Bengio, Y., Courville, A., and Vincent, P., \enquote{Representation learning: A review and new perspectives,} \emph{IEEE Transactions on Pattern Analysis and Machine Intelligence}, Vol.~35, No.~8, 2013, pp. 1798--1828.

\bibitem[{Chen et~al.(2020)Chen, Chiu, and Fuge}]{chen2020airfoil}
Chen, W., Chiu, K., and Fuge, M.~D., \enquote{Airfoil design parameterization and optimization using b{\'e}zier generative adversarial networks,} \emph{AIAA Journal}, Vol.~58, No.~11, 2020, pp. 4723--4735.

\bibitem[{Goodfellow et~al.(2020)Goodfellow, Pouget-Abadie, Mirza, Xu, Warde-Farley, Ozair, Courville, and Bengio}]{goodfellow2020generative}
Goodfellow, I., Pouget-Abadie, J., Mirza, M., Xu, B., Warde-Farley, D., Ozair, S., Courville, A., and Bengio, Y., \enquote{Generative adversarial networks,} \emph{Communications of the ACM}, Vol.~63, No.~11, 2020, pp. 139--144.

\bibitem[{Kingma and Welling(2013)}]{kingma2013auto}
Kingma, D.~P., and Welling, M., \enquote{Auto-encoding variational bayes,} \emph{arXiv preprint arXiv:1312.6114}, 2013.

\bibitem[{Arjovsky et~al.(2017)Arjovsky, Chintala, and Bottou}]{arjovsky2017wasserstein}
Arjovsky, M., Chintala, S., and Bottou, L., \enquote{Wasserstein generative adversarial networks,} \emph{International Conference on Machine Learning}, PMLR, Sydney, Austrailia, 2017, pp. 214--223.

\bibitem[{Salimans et~al.(2016)Salimans, Goodfellow, Zaremba, Cheung, Radford, and Chen}]{salimans2016improved}
Salimans, T., Goodfellow, I., Zaremba, W., Cheung, V., Radford, A., and Chen, X., \enquote{Improved techniques for training gans,} \emph{Advances in Neural Information Processing Systems}, Vol.~29, 2016.

\bibitem[{Razavi et~al.(2019)Razavi, Van~den Oord, and Vinyals}]{razavi2019generating}
Razavi, A., Van~den Oord, A., and Vinyals, O., \enquote{Generating diverse high-fidelity images with vq-vae-2,} \emph{Advances in Neural Information Processing Systems}, Vol.~32, 2019.

\bibitem[{Yilmaz and German(2020)}]{yilmaz2020conditional}
Yilmaz, E., and German, B., \enquote{Conditional generative adversarial network framework for airfoil inverse design,} \emph{AIAA Aviation 2020 Forum}, Online, 2020.

\bibitem[{Achour et~al.(2020)Achour, Sung, Pinon-Fischer, and Mavris}]{achour2020development}
Achour, G., Sung, W.~J., Pinon-Fischer, O.~J., and Mavris, D.~N., \enquote{Development of a conditional generative adversarial network for airfoil shape optimization,} \emph{AIAA Scitech 2020 Forum}, Orlando, Florida, 2020.

\bibitem[{Yonekura and Suzuki(2021)}]{yonekura2021data}
Yonekura, K., and Suzuki, K., \enquote{Data-driven design exploration method using conditional variational autoencoder for airfoil design,} \emph{Structural and Multidisciplinary Optimization}, Vol.~64, No.~2, 2021, pp. 613--624.

\bibitem[{Yang et~al.(2023)Yang, Lee, and Yee}]{yang2023inverse}
Yang, S., Lee, S., and Yee, K., \enquote{Inverse design optimization framework via a two-step deep learning approach: application to a wind turbine airfoil,} \emph{Engineering with Computers}, Vol.~39, No.~3, 2023, pp. 2239--2255.

\bibitem[{Du et~al.(2020)Du, He, and Martins}]{du2020b}
Du, X., He, P., and Martins, J.~R., \enquote{A B-spline-based generative adversarial network model for fast interactive airfoil aerodynamic optimization,} \emph{AIAA Scitech 2020 Forum}, Orlando, Florida, 2020.

\bibitem[{Wang et~al.(2023{\natexlab{a}})Wang, Shimada, and Farimani}]{wang2023airfoil}
Wang, Y., Shimada, K., and Farimani, A.~B., \enquote{Airfoil GAN: Encoding and synthesizing airfoils for aerodynamic shape optimization,} \emph{Journal of Computational Design and Engineering}, 2023{\natexlab{a}}.

\bibitem[{Kou et~al.(2023)Kou, Botero-Bol{\'\i}var, Ballano, Marino, de~Santana, Valero, and Ferrer}]{kou2023aeroacoustic}
Kou, J., Botero-Bol{\'\i}var, L., Ballano, R., Marino, O., de~Santana, L., Valero, E., and Ferrer, E., \enquote{Aeroacoustic airfoil shape optimization enhanced by autoencoders,} \emph{Expert Systems with Applications}, Vol. 217, 2023.

\bibitem[{Savitzky and Golay(1964)}]{savitzky1964smoothing}
Savitzky, A., and Golay, M.~J., \enquote{Smoothing and differentiation of data by simplified least squares procedures.} \emph{Analytical chemistry}, Vol.~36, No.~8, 1964, pp. 1627--1639.

\bibitem[{Wada et~al.(2023)Wada, Suzuki, and Yonekura}]{wada2023physics}
Wada, K., Suzuki, K., and Yonekura, K., \enquote{Physics-guided training of GAN to improve accuracy in airfoil design synthesis,} \emph{arXiv preprint arXiv:2308.10038}, 2023.

\bibitem[{Lin et~al.(2022)Lin, Zhang, Xie, Shi, Xu, and Duan}]{lin2022cst}
Lin, J., Zhang, C., Xie, X., Shi, X., Xu, X., and Duan, Y., \enquote{CST-GANs: A Generative adversarial network Based on CST parameterization for the generation of smooth airfoils,} \emph{2022 IEEE International Conference on Unmanned Systems (ICUS)}, IEEE, Guangzhou, China, 2022, pp. 600--605.

\bibitem[{Du et~al.(2022)Du, Liu, Yang, Li, Zhang, and Xie}]{du2022airfoil}
Du, Q., Liu, T., Yang, L., Li, L., Zhang, D., and Xie, Y., \enquote{Airfoil design and surrogate modeling for performance prediction based on deep learning method,} \emph{Physics of Fluids}, Vol.~34, No.~1, 2022.

\bibitem[{Higgins et~al.(2016)Higgins, Matthey, Pal, Burgess, Glorot, Botvinick, Mohamed, and Lerchner}]{higgins2016beta}
Higgins, I., Matthey, L., Pal, A., Burgess, C., Glorot, X., Botvinick, M., Mohamed, S., and Lerchner, A., \enquote{beta-vae: Learning basic visual concepts with a constrained variational framework,} \emph{International Conference on Learning Representations}, Toulon, France, 2016.

\bibitem[{Burgess et~al.(2018)Burgess, Higgins, Pal, Matthey, Watters, Desjardins, and Lerchner}]{burgess2018understanding}
Burgess, C.~P., Higgins, I., Pal, A., Matthey, L., Watters, N., Desjardins, G., and Lerchner, A., \enquote{Understanding disentangling in $\beta$-VAE,} \emph{arXiv preprint arXiv:1804.03599}, 2018.

\bibitem[{Rolinek et~al.(2019)Rolinek, Zietlow, and Martius}]{rolinek2019variational}
Rolinek, M., Zietlow, D., and Martius, G., \enquote{Variational autoencoders pursue pca directions (by accident),} \emph{Proceedings of the IEEE/CVF Conference on Computer Vision and Pattern Recognition}, Long Beach, California, 2019, pp. 12406--12415.

\bibitem[{Burda et~al.(2015)Burda, Grosse, and Salakhutdinov}]{burda2015importance}
Burda, Y., Grosse, R., and Salakhutdinov, R., \enquote{Importance weighted autoencoders,} \emph{arXiv preprint arXiv:1509.00519}, 2015.

\bibitem[{S{\o}nderby et~al.(2016)S{\o}nderby, Raiko, Maal{\o}e, S{\o}nderby, and Winther}]{sonderby2016ladder}
S{\o}nderby, C.~K., Raiko, T., Maal{\o}e, L., S{\o}nderby, S.~K., and Winther, O., \enquote{Ladder variational autoencoders,} \emph{Advances in Neural Information Processing systems}, Vol.~29, 2016.

\bibitem[{Takeishi and Kalousis(2021)}]{takeishi2021physics}
Takeishi, N., and Kalousis, A., \enquote{Physics-integrated variational autoencoders for robust and interpretable generative modeling,} \emph{Advances in Neural Information Processing Systems}, Vol.~34, 2021, pp. 14809--14821.

\bibitem[{Dai and Wipf(2019)}]{dai2019diagnosing}
Dai, B., and Wipf, D., \enquote{Diagnosing and enhancing VAE models,} \emph{arXiv preprint arXiv:1903.05789}, 2019.

\bibitem[{Eivazi et~al.(2022)Eivazi, Le~Clainche, Hoyas, and Vinuesa}]{eivazi2022towards}
Eivazi, H., Le~Clainche, S., Hoyas, S., and Vinuesa, R., \enquote{Towards extraction of orthogonal and parsimonious non-linear modes from turbulent flows,} \emph{Expert Systems with Applications}, Vol. 202, 2022.

\bibitem[{Li et~al.(2022)Li, Zhang, and Chen}]{li2022physically}
Li, R., Zhang, Y., and Chen, H., \enquote{Physically Interpretable Feature Learning of Supercritical Airfoils Based on Variational Autoencoders,} \emph{AIAA Journal}, Vol.~60, No.~11, 2022, pp. 6168--6182.

\bibitem[{Kang et~al.(2022)Kang, Yang, and Yee}]{kang2022physics}
Kang, Y.-E., Yang, S., and Yee, K., \enquote{Physics-aware reduced-order modeling of transonic flow via $\beta$-variational autoencoder,} \emph{Physics of Fluids}, Vol.~34, No.~7, 2022.

\bibitem[{Wang et~al.(2023{\natexlab{b}})Wang, Xie, Zhang, and Xu}]{wang2023physics}
Wang, J., Xie, H., Zhang, M., and Xu, H., \enquote{Physics-assisted reduced-order modeling for identifying dominant features of transonic buffet,} \emph{Physics of Fluids}, Vol.~35, No.~6, 2023{\natexlab{b}}.

\bibitem[{O'Shea and Nash(2015)}]{o2015introduction}
O'Shea, K., and Nash, R., \enquote{An introduction to convolutional neural networks,} \emph{arXiv preprint arXiv:1511.08458}, 2015.

\bibitem[{Drela(1989)}]{drela1989xfoil}
Drela, M., \enquote{XFOIL: An analysis and design system for low Reynolds number airfoils,} \emph{Low Reynolds Number Aerodynamics: Proceedings of the Conference Notre Dame}, Springer, Berlin, Heidelberg, 1989, pp. 1--12.

\end{thebibliography}

\end{document}